\documentclass{article}

\usepackage{arxiv}

\usepackage[utf8]{inputenc} 
\usepackage[T1]{fontenc}    
\usepackage{hyperref}       
\usepackage{url}            
\usepackage{booktabs}       
\usepackage{amsfonts}       
\usepackage{nicefrac}       
\usepackage{microtype}      
\usepackage{lipsum}		
\usepackage{graphicx}
\usepackage{natbib}
\usepackage{doi}

\usepackage[utf8]{inputenc} 
\usepackage[T1]{fontenc}    
\usepackage{hyperref}       
\usepackage{url}            
\usepackage{booktabs}       
\usepackage{nicefrac}       
\usepackage{microtype}      
\usepackage[table]{xcolor}         
\definecolor{lightblue}{rgb}{0.85,0.85,1.0}
\definecolor{lightgray}{rgb}{0.95,0.95,0.95}
\usepackage{subcaption}

\usepackage{amsmath}
\usepackage{mathtools}
\usepackage{amsthm}

\theoremstyle{plain}

\theoremstyle{definition}

\theoremstyle{remark}

\usepackage{enumitem}
\setlist{nosep}
\setlist[itemize]{leftmargin=20pt}

\usepackage{graphicx}
\usepackage{bbm}
\usepackage{makecell}
\usepackage{multirow, multicol}
\usepackage{wrapfig}
\usepackage{utfsym}
\usepackage{threeparttable}

\definecolor{lightyellow}{HTML}{FFFFE0}
\definecolor{lightpurple}{HTML}{D8BFD8}

\usepackage{listings}
\usepackage{xcolor}
\usepackage[most]{tcolorbox}

\lstdefinestyle{prompt}{
  basicstyle=\ttfamily\small,
  columns=fullflexible,
  breaklines=true,
  breakatwhitespace=true,
  keepspaces=true,
  showstringspaces=false,
  frame=none,
  aboveskip=0pt,
  belowskip=0pt,
  language={},
}

\title{CLUBench: A Clustering Benchmark}

\author{ Feng Xiao\thanks{Both authors contributed equally to this research.} \\
	The Chinese University of Hong Kong (Shenzhen)\\
	\texttt{fengxiao1@link.cuhk.edu.cn} \\
	\And
	Dazhi Fu$^*$ \\
	The Chinese University of Hong Kong (Shenzhen)\\
	\texttt{dazhifu@link.cuhk.edu.cn} \\
	\AND
    Chris Ding \\
	The Chinese University of Hong Kong (Shenzhen)\\
	\texttt{chrisding@cuhk.edu.cn}
	\And
	Jicong Fan\thanks{Corresponding author.}\\
	The Chinese University of Hong Kong (Shenzhen)\\
	\texttt{fanjicong@cuhk.edu.cn}
}

\renewcommand{\shorttitle}{\textit{arXiv} Template}

\hypersetup{
pdftitle={A template for the arxiv style},
pdfsubject={q-bio.NC, q-bio.QM},
pdfauthor={David S.~Hippocampus, Elias D.~Striatum},
pdfkeywords={First keyword, Second keyword, More},
}

\begin{document}
\maketitle

\begin{abstract}
Clustering is a fundamental problem in data science with a long-standing research history, yielding numerous insightful algorithms.
Despite this progress, a systematic and large-scale empirical evaluation that jointly considers conventional algorithms, deep learning-based methods, and recent foundation model-based clustering remains largely absent, leading to limited guidance on algorithm selection and deployment. To address this gap, we introduce CLUBench, a comprehensive clustering benchmark comprising 24 algorithms of diverse principles evaluated on 131 datasets across tabular, text, and image data, involving 178,815 experiments. Importantly, our analyses of 
(i) the impact of hyperparameter tuning,
(ii) the impact of data types and characteristics,
(iii) the impact of pretrained embeddings,
(iv) large language model-based clustering,
(v) the similarity of algorithms, and
(vi) the low-rank structures of performance matrices,
yield meaningful insights and promising pathways for clustering research. 
For instance, our study reveals that: 1) All evaluated deep clustering methods do not exhibit a significant advantage compared with the top-performing conventional clustering algorithms (e.g., KMeans, SpeClu) in terms of average performance; 2) For image and text clustering tasks, combining pretrained embeddings with conventional clustering algorithms (e.g., KMeans, SpeClu) offers effective and efficient clustering; 3) Clustering remains a challenging and nontrivial problem, even in the era of increasingly dominant foundation models.
Moreover, we propose to use the low-rank structure in cross-model performance matrices to efficiently approximate the overall performance evaluation in practical applications. 
We further demonstrate the feasibility of model selection based on the performance matrices across all hyperparameter configurations. 
Finally, we provide an easy-to-use toolbox by encapsulating the source code from the official code repository into a unified framework, accompanied by detailed instructions. All benchmark datasets, overall experimental results, and the toolbox are fully open-sourced and available at \url{https://github.com/xiaofeng-github/CLUBench}.

\end{abstract}


\section{Introduction}

We are living in a world full of data, which serves as an approximate reflection of the physical reality. One of the basic means of mining these data is to classify samples into a set of categories for more granular investigation. As a result, clustering tasks are ubiquitous in the real world and cluster analysis becomes a fundamental technique used in various fields, including pattern recognition, information retrieval, bioinformatics, data compression, etc. Since the 1960s, there have been systematic studies~\citep{forgy1965cluster, mcqueen1967some} concerning the clustering problem. Over the years, numerous clustering algorithms have been developed based on different observations or assumptions. Following the de facto standard taxonomy~\citep{xu2015comprehensive, yin2024rapid}, conventional clustering algorithms can be organized into six categories. The first category is partition-based clustering with classic algorithms like K-means~\citep{mcqueen1967some}, K-medoids~\citep{park2009simple} and CLARANS~\citep{ng2002clarans}. The second category is hierarchical clustering~\citep{zhang1996birch, 10.1145/3637528.3671888} which constructs the hierarchical relationship among data. The third category is density-based clustering~\citep{ester1996density, ankerst1999optics, comaniciu2002mean}. The fourth category is subspace clustering~\citep{elhamifar2013sparse,chen2020stochastic}, which discovers clusters that exist in different (possibly overlapping) low-dimensional subspaces of data. The fifth category is model-based clustering, where each cluster is assigned a particular model (e.g. GMM~\citep{rasmussen1999infinite}) and the core idea is to find the best fit between clusters and models. The last category is graph-based clustering, which treats data as graphs and applies spectral graph theory to identify communities, such as spectral clustering~\cite{von2007tutorial} and AutoSC\cite{fan2022simple}. With the progress of deep learning techniques and especially deep unsupervised learning, many deep neural network-based clustering (DC) methods~\citep{dec,idec,dscn,pica,cc,edesc,DMICC,DIVC,LFSS} have been proposed in the past few years. These DC methods exhibit marked superiority when dealing with complex and high-dimensional data compared with conventional clustering algorithms. 

While the proliferation of clustering techniques has greatly enriched the methodological landscape, it has also intensified the challenge of method selection and application, as different algorithms often exhibit highly inconsistent performance across datasets, domains, and experimental settings. In response to the rapid proliferation and diversity of clustering methods, a number of survey and review articles~\citep{jain1999data, xu2005survey, berkhin2006survey, xu2015comprehensive, min2018survey, aljalbout2018clustering, nutakki2018introduction, liu2022survey} have been proposed to systematically organize and compare existing techniques. More recently, several comprehensive reviews~\citep{yin2024rapid, wei2024overview, zhou2024comprehensive, ren2024deep} have further highlighted the breadth, heterogeneity, and continual expansion of the clustering literature. Despite their considerable value in providing taxonomies, conceptual summaries, and qualitative comparisons, these surveys generally lack comprehensive, quantitatively grounded evaluations and analyses that can faithfully characterize the empirical behavior of modern clustering algorithms. In addition, the benchmark literature for clustering remains limited and incomplete. Although a few works~\citep{javed2020benchmark, leiber2023benchmarking, zhou2024comprehensive, wei2024overview} have attempted to fill this gap, they have limitations:

\begin{table*}[!ht]
    \caption{Comparison between CLUBench and existing clustering surveys/benchmarks. `DC' and `FM' mean `Deep Clustering' and `Foundation Models', respectively. The $^*$ marks the clustering surveys with experimental evaluations.}
    \label{tab-benchmark-com}
    \centering
    \begin{threeparttable}
    \resizebox{\textwidth}{!}{
    \begin{tabular}{l|c|c|c|c|c|c|c|c|c|c|c|c|c|c|c|c}
    \toprule
    \multirow{2}{*}{\textbf{Benchmark}} & \multicolumn{3}{c|}{\textbf{Coverage}} & \multicolumn{8}{c|}{\textbf{Algorithm Type (-based)}} & \multicolumn{3}{c|}{\textbf{Data Type}} & \multicolumn{2}{c}{\textbf{Resource Integration}}\\
    & \# Datasets & \# Algo. & \# Metrics & Partition & Hierarchy & Density & Model & Subspace & Graph & DC & FM & Tabular & Image & Sequence & Data & Toolbox\\
    \midrule
    \cite{xu2005survey} (2005)&2&15&1 & \usym{1F5F8} & \usym{1F5F8} & \usym{1F5F8} & \usym{1F5F8} & \usym{2717}  & \usym{1F5F8} &\usym{1F5F8} & \usym{2717} &  \usym{1F5F8} & \usym{2717} & \usym{2717} &   \usym{2717} &  \usym{2717}\\
     \cite{thrun2020clustering} (2020)&12 &0&0 & \usym{2717} & \usym{2717} & \usym{2717} & \usym{2717} & \usym{2717}  & \usym{2717}  & \usym{2717} & \usym{2717} &  \usym{1F5F8} & \usym{2717} & \usym{2717} &  \usym{1F5F8} & \usym{1F5F8}~\tnote{1}\\
    \cite{javed2020benchmark} (2020)& 112 & 8 & 1 &  \usym{1F5F8} & \usym{1F5F8} & \usym{1F5F8} & \usym{2717} & \usym{2717} &  \usym{2717} &  \usym{2717} & \usym{2717} & \usym{2717} & \usym{2717} & \usym{1F5F8} & \usym{2717} & \usym{1F5F8}~\tnote{2}\\
    \cite{shand2021hawks} (2021)&6 &5&1 & \usym{1F5F8} & \usym{1F5F8} & \usym{2717} & \usym{1F5F8} & \usym{2717}  & \usym{2717}& \usym{2717} & \usym{2717} &  \usym{1F5F8} & \usym{2717} & \usym{2717} &  \usym{1F5F8} & \usym{1F5F8}~\tnote{3}\\
    \cite{leiber2023benchmarking} (2023)& 8 & 6 & 3 & \usym{2717} & \usym{2717} & \usym{2717} & \usym{2717} & \usym{2717} & \usym{2717} & \usym{1F5F8}& \usym{2717} & \usym{2717} & \usym{1F5F8} & \usym{2717} &  \usym{2717} & \usym{1F5F8}~\tnote{4}\\
    \cite{zhou2024comprehensive}$^*$ (2024) & 4 & 2 & 1 & \usym{2717} & \usym{2717} & \usym{2717} & \usym{2717} & \usym{2717} & \usym{2717} &  \usym{1F5F8} & \usym{2717} &  \usym{2717} & \usym{1F5F8} & \usym{2717} &  \usym{1F5F8} & \usym{1F5F8}~\tnote{5}\\
    \cite{wei2024overview}$^*$ (2024)& 12 & 26 & 3 & \usym{2717} & \usym{2717} & \usym{2717} & \usym{2717} & \usym{2717} & \usym{2717} & \usym{1F5F8}& \usym{2717} &  \usym{1F5F8} &  \usym{1F5F8} &  \usym{1F5F8} & \usym{2717} & \usym{2717}\\
    \cite{xu2025scclubench} (2025) & 36 & 14 & 3 & \usym{2717} & \usym{1F5F8} & \usym{2717} & \usym{2717} & \usym{2717} & \usym{1F5F8}& \usym{1F5F8}& \usym{1F5F8} &  \usym{1F5F8} &  \usym{2717} &  \usym{2717} & \usym{2717} & \usym{2717}\\
    \midrule
    CLUBench (ours) & 131 & 24 & 3 & \usym{1F5F8} & \usym{1F5F8} & \usym{1F5F8} & \usym{1F5F8} & \usym{1F5F8} & \usym{1F5F8} & \usym{1F5F8} & \usym{1F5F8} & \usym{1F5F8} & \usym{1F5F8} & \usym{1F5F8} & \usym{1F5F8} & \usym{1F5F8}~\tnote{6} \\
    \bottomrule
    \end{tabular}
    }
 \begin{tablenotes}[flushleft]
\tiny
\setlength{\itemsep}{0pt}
\item 
$[1]$ \url{https://data.mendeley.com/datasets/vsxvgc4rwy/1}
$[2]$ \url{https://github.com/ali-javed/clusteringBenchmark}
$[3]$ \url{https://github.com/sea-shunned/hawks}
\item 
$[4]$ \url{https://github.com/collinleiber/ClustPy}
$[5]$ \url{https://github.com/zhoushengisnoob/OpenDeepClustering} 
$[6]$ \url{https://github.com/xiaofeng-github/CLUBench}
\end{tablenotes}
    \end{threeparttable}
\end{table*}

\begin{enumerate}
    \item The evaluated benchmark datasets and clustering algorithms are limited in both quantity and diversity, which can easily lead to biased evaluations and conclusions.

    \item The evaluations typically focus on either conventional clustering algorithms or deep learning-based methods, lacking a joint and unified comparison across paradigms.

    \item A comparison between state-of-the-art clustering methods and foundation model-based clustering is largely missing, despite the growing dominance of foundation models.

    \item There remains no convenient, unified toolbox for clustering methods, particularly deep clustering.

\end{enumerate}

In this benchmark, we attempt to address these limitations. Based on the existing work~\citep{jeon2025measuring} and publicly available dataset archives, we collect 131 datasets from diverse real-world domains, covering three data modalities: tabular, text, and image. These datasets are evaluated on 24 clustering algorithms, and we provide complete experimental results, systematic comparisons, and in-depth analyses. Notably, we conduct a unified comparison between state-of-the-art clustering baselines and foundation model-based clustering strategies, enabling an examination of their performance gaps and offering insights to guide future clustering research. In addition, we integrate the source code, particularly the implementations of deep learning methods, into a unified framework that provides a consistent and convenient interface. The main contributions of this benchmark are summarized as follows:

\begin{itemize}
    
    \item This benchmark provides a quantitative evaluation on 131 datasets across three modalities (tabular, text, and image), where 24 clustering methods, including conventional and deep clustering algorithms, are considered. In addition, a comparison between state-of-the-art clustering baselines and foundation model-based strategies is provided. 
    
    \item From an application perspective, this benchmark supports effective algorithm selection through preference analysis, enables efficient performance evaluation via low-rank analysis, and offers a practical strategy for model selection—covering both algorithm choice and hyperparameter configuration.

    \item From a research perspective, this work introduces a challenging benchmark for clustering tasks, as the best-performing baselines remain moderate in performance, and identifies several valuable phenomena and promising directions for future clustering research.

    \item This benchmark provides a unified toolbox to support practical deployment and further research in clustering, which can be easily extended to new datasets and algorithms.

\end{itemize}


\section{Related Work}
\subsection{Clustering Algorithms}

In general, clustering algorithms can be broadly categorized into two groups based on whether neural networks are employed: conventional clustering algorithms \citep{kmeans++, ng2001spectral, comaniciu2002mean,elhamifar2013sparse,chen2020stochastic,fan2021large,liu2012robust,fan2022simple}, and deep clustering algorithms \citep{idec,dscn,caron2018deep,han2019learning,asano2019self,pica,edesc,DIVC,zhang2024p,LFSS,VDE,JULE,DAC,DCC,DSEC,DDCA,niu2022spice,nie2023fast,do2021clustering,peng2025provable,guo2022hcsc,qian2023stable,li2023image,liu2024interactive,shen2021you, zhu2025hierarchical, li2025you}. Due to space limitations, a more detailed review and finer-grained categorization of these algorithms are provided in \text{Appendix}~\ref{app-CA}.

\subsection{Existing Clustering Benchmarks and Reviews with Experimental Evaluation}

As discussed above, numerous conventional and deep clustering algorithms have been developed over the past decades. A systematic evaluation of these methods on diverse real-world datasets, together with an easy-to-use implementation toolbox, is therefore essential for advancing the field. Prior studies~\citep{jain1999data, xu2005survey, berkhin2006survey, omran2007overview, von2010clustering, murtagh2012algorithms, xu2015comprehensive, min2018survey, aljalbout2018clustering, nutakki2018introduction, javed2020benchmark, liu2022survey, yin2024rapid, ren2024deep, zhou2024comprehensive, wei2024overview, wehrli2023german, o2025openclustered} have reviewed clustering algorithms from various perspectives.
For example, \citet{jain1999data} reviews partition-based and hierarchical methods, discussing their theoretical foundations and applications.  \citet{leiber2023benchmarking} focuses on deep clustering for image data and introduces Clustpy, a unified benchmarking framework for fair comparison. \citet{zhou2024comprehensive} decomposes deep clustering methods into representation learning and clustering components and proposes a taxonomy based on their interaction.  More details about prior studies are provided in Appendix~\ref{app-bench-related}. Collectively, these studies offer valuable insights into the development and application of clustering techniques. However, despite their contributions, several limitations remain: evaluations are typically conducted on a limited set of algorithms and datasets; joint comparisons across traditional, deep learning, and foundation model-based methods are largely missing; and no unified, easy-to-use toolbox is available to support reproducible benchmarking. 
As summarized in Table~\ref{tab-benchmark-com}, compared with existing studies, CLUBench provides a substantially larger experimental scale, more comprehensive and in-depth analyses, and an easy-to-use toolbox.

\section{CLUBench}

Clustering encompasses several distinct methodological topics.
This benchmark is specifically centered on classical \textbf{hard clustering}, and does not include \textcolor{gray!70}{Soft Clustering \cite{kumar2007soft,peters2013soft,ferraro2020soft}}, \textcolor{gray!70}{Multi-view Clustering~\citep{Kang_Zhou_Zhao_Shao_Han_Xu_2020,10108535,chen2023deep}}, \textcolor{gray!70}{Online Clustering~\citep{beringer2006online, barbakh2008online, li2022twin}}, \textcolor{gray!70}{Semi-Clustering~\cite{basu2004probabilistic, bair2013semi, cai2023review}}, \textcolor{gray!70}{Graph Clustering (clustering object:graph)~\citep{tian2014learning,liu2023simple,10.1145/3690624.3709207}}, \textcolor{gray!70}{Distributed Clustering~\cite{hai2012survey,qiao2023federated}} or other possible topics.
Following the description in prior work~\citep{hansen1997cluster}, a concise mathematical formulation that considers only the inputs and final outputs of algorithms is presented in Section~\ref{hard-clustering}. In addition, to facilitate navigation of the benchmark, we provide a guidance map in Section~\ref{benchmark-instruction}, which offers a high-level overview of the research scope in this benchmark.
    
\subsection{Problem Description}
\label{hard-clustering}

    Given a dataset $\mathcal{D}=\{\mathbf{x}_1, \mathbf{x}_2, \cdots, \mathbf{x}_n\}$ with $n$ samples where each  $\mathbf{x}_i \in \mathcal{X}$. In CLUBench, a clustering algorithm $f$ aims to seek a $K$-partition of $\mathcal{D}$, $C=\{C_1, C_2, \cdots, C_K\} (K<n)$, such that
    \begin{enumerate}
        \item $C_i \neq \emptyset, i=1, \dots, K$;
        \item $\bigcup_{i=1}^{K} C_i = \mathcal{D}$;
        \item $C_i \cap C_j = \emptyset, i, j=1,\dots, K$ and $i \neq j$.
    \end{enumerate}
where (3) indicates that all algorithms belong to hard clustering. $K$ must be determined prior to learning for some algorithms, whereas for others it is determined during the learning process. 

\subsection{Roadmap of CLUBench}
\label{benchmark-instruction}

\begin{enumerate}
    
    \item \textbf{Algorithms.} In CLUBench, we assemble a comprehensive collection of 24 clustering algorithms, comprising 14 conventional clustering (CC) algorithms and 10 deep clustering (DC) methods. Following the categorization methodology of prior clustering reviews~\cite{xu2005survey, xu2015comprehensive, yin2024rapid} on conventional algorithms, six distinct categories are covered in CLUBench:
    \begin{itemize}
        \item \textbf{Partition-based}: KMeans~\cite{mcqueen1967some}, KernelKMeans~\cite{dhillon2004kernel}, k-PC~\cite{agarwal2004k}.

        \item \textbf{Hierarchical}: AggClu (Agglomerative Clustering)~\cite{johnson1967hierarchical}, BIRCH~\cite{zhang1996birch}.

        \item \textbf{Density-based}: DBSCAN~\cite{ester1996density}, MeanShift~\cite{comaniciu2002mean}.

        \item \textbf{Model-based}: GMM~\cite{rasmussen1999infinite}.

        \item \textbf{Subspace-based}: SSC~\citep{elhamifar2013sparse}, S$^3$COMP-C~\citep{chen2020stochastic}, k-FSC~\citep{fan2021large}, LRR~\citep{liu2012robust}.

        \item \textbf{Graph-based}: SpeClu~\citep{von2007tutorial}, AutoSC\cite{fan2022simple}.
    \end{itemize}
    For deep clustering, the 10 state-of-the-art clustering methods, published from 2016 to 2025, are collected based on their distinct technical paradigms:
    \begin{itemize}
        \item \textbf{Autoencoder-based}: DEC~\cite{dec}, IDEC~\cite{idec}.

        \item \textbf{Deep Subspace Clustering}: DSCN~\cite{dscn}, EDESC~\cite{edesc}.

        \item \textbf{Contrastive Learning-based}: ConClu (Contrastive Clustering)~\citep{cc}, DMICC~\cite{DMICC}.

        \item \textbf{Information Theory-based}: PICA~\cite{pica}, DIVC~\cite{DIVC}.


        \item \textbf{Self-Supervision-based}: P$^2$OT~\cite{zhang2024p}, LFSS~\cite{LFSS}.

    \end{itemize}
    
    More details about time and space complexity and hyperparameter configurations (HPC) of the evaluated clustering algorithms are provided in Appendix~\ref{appe-algorithms}.

    \item \textbf{Datasets.} Based on previous works~\citep{wei2024overview, zhou2024comprehensive, jeon2025measuring} and publicly available dataset archive\footnote{https://www.openml.org/search?type=data\&status=active}, we collect 131 benchmark datasets for clustering evaluation. A statistical summary is provided in Table~\ref{datasets-summary}.
    \begin{table}[!ht]
        \caption{Dataset summary. $r_{mm}:=\frac{\#~\text{samples in minimal cluster}}{\#~\text{samples in maximal cluster}} \in (0, 1]$ measures the imbalance ratio between clusters.}
        \label{datasets-summary}
        \centering
        \resizebox{0.6\textwidth}{!}{
        \begin{tabular}{|c|c|c|}
        \hline
        \textbf{Type} & \textbf{\# Samples} & \# \textbf{Dimension} \\
        \hline
        tabular, text, image & 61 -  10,000 & 2 - 27,648 \\
        \hline
        \textbf{Domain} & \textbf{\# Clusters} &\#  \textbf{$r_{\text{mm}} \in (0.0, 1.0]$} \\
        \hline
        \makecell{medical, biology, finance, \\agriculture, astronomy, industry} & 2 - 40 & 0.06 - 1.00 \\
        \hline
        \end{tabular}}
    \end{table}
    Detailed descriptions of all datasets are provided in Appendix~\ref{appe-datasets}. It is noteworthy that the majority of datasets in CLUBench are tabular. This preference stems from the diversity of tabular data and the capacity to abstract away domain-specific intricacies and thereby present the pure clustering problem of discovering groups based on distance, density, or distribution within a D-dimensional feature space. For completeness, CLUBench also incorporates representative image and text datasets.

    \item \textbf{Overall Performance Comparison.} 
    The aggregate performance of the clustering algorithms, evaluated by the average ACC, Normalized Mutual Information (NMI), and Adjusted Rand Index (ARI) across the 131 datasets, is presented in Section~\ref{analysis-1}. The analysis shows that deep clustering does not significantly outperform conventional algorithms like KMeans and spectral clustering (SpeClu). Complete results for each algorithm and dataset are in Appendix~\ref{appe-complete-all-performance}.

    \item \textbf{Foundation Model-based Clustering Analysis.}
    Motivated by the rapid advancement of foundation models (FM), in Section~\ref{analysis-fm}, we incorporate a comprehensive set of FM-based comparisons into our study. For image clustering, we compare several top-performing baselines (identified in Section~\ref{analysis-1}) operating on feature embeddings extracted from pretrained models with state-of-the-art image-based deep clustering methods. For text clustering, we evaluate clustering algorithms on embeddings derived from large language models (LLMs). For tabular data, we adopt an in-context learning paradigm with LLMs to perform clustering.
    
    \item \textbf{Similarity Analysis of Algorithms and Datasets.}
    Based on the obtained performance results (ACC, NMI, ARI), we can construct a unique performance vector for each clustering algorithm and each dataset, respectively. Leveraging these vectors, Section~\ref{analysis-3} investigates performance-based similarities among algorithms as well as among datasets.

    \item \textbf{Algorithm Preference Analysis.} 
    To investigate the potential preferences of the clustering algorithms, we group the datasets according to different criteria: (i) data type (image, text, tabular, bioinformatics); (ii) feature dimensionality (low, mid, high); (iii) imbalance ratio (low, mid, high). In Section~\ref{analysis-2}, the analysis indicates that the algorithm performance rankings are highly context-dependent and algorithms with strong overall performance may still be suboptimal under specific dataset conditions.

    \item \textbf{Low-Rank Analysis of the Performance Matrix.}
    By tuning hyperparameter configurations (HPCs) for each algorithm, we construct performance matrices under different evaluation metrics (ACC, NMI, and ARI). Specifically, when each algorithm is evaluated under $h$ HPCs, we obtain performance matrices $\mathbf{\hat{P}}_{\text{acc}}, \mathbf{\hat{P}}_{\text{nmi}}, \mathbf{\hat{P}}_{\text{ari}} \in \mathbb{R}^{131 \times (24h)}$, corresponding to 131 datasets and 24 algorithms. Based on these matrices, Section~\ref{analysis-4} investigates their low-rank structure and further explores matrix completion to assess the feasibility and effectiveness of fast performance prediction from incomplete performance observations.

    \item \textbf{Model Selection based on Performance Matrices.} 
    The empirical evaluations reveal that the best-performing algorithms or HPCs for different datasets vary due to differences in their structure and meta-features, underscoring the necessity of effective model selection in clustering research. Accordingly, in Section~\ref{analysis-ms}, we seek to learn an effective mapping $f$ between dataset meta-features and the corresponding performance matrices $\mathbf{\hat{P}}_{\text{acc}}, \mathbf{\hat{P}}_{\text{nmi}}, \mathbf{\hat{P}}_{\text{ari}}$.

    \item \textbf{Toolbox.}
    To facilitate both practical application and further research on the wide range of clustering methods, we provide an easy-to-use toolbox encapsulated in Python and fully compatible with \texttt{scikit-learn}~\footnote{\url{https://scikit-learn.org/stable/index.html}}. The toolbox follows a unified and intuitive usage paradigm across different algorithms. An example of using \texttt{DEC} is shown below:
\begin{lstlisting}[basicstyle=\ttfamily\tiny, frame=single, breaklines=true, xleftmargin=5pt, xrightmargin=5pt]
CM = DEC(**hpc) # hpc: dict of hyperparameter configurations.
CM.fit_predict(X) # X: data matrix of shape (n_samples, dim).
CM.labels # predicted labels.
CM.time # runtime.
acc, nmi, ari = CM.evaluation(Y)  # Y: true labels.
\end{lstlisting}
New datasets and clustering algorithms can be easily integrated into the toolbox. Further details are available in the~\href{https://github.com/xiaofeng-github/CLUBench}{code repository}.


    \end{enumerate}
 
\section{Experiment Results and Analysis}

\subsection{Overall Performance Comparison}
\label{analysis-1}

\begin{table*}[!ht]

\centering
\caption{Average performance (ACC, NMI, ARI) across all datasets. The best performances within `CC' and `DC' are in \textcolor{orange}{bold}. } 
\label{tab-all-avg-p}
\resizebox{\textwidth}{!}{
\begin{tabular}{l*{8}{|l}}
\toprule
\textbf{Algorithms} 
  & \multicolumn{1}{c|}{KMeans (CC)}
  & \multicolumn{1}{c|}{KerKMeans (CC)}
  & \multicolumn{1}{c|}{AggClu (CC)}
  & \multicolumn{1}{c|}{DBSCAN (CC)}
  & \multicolumn{1}{c|}{BIRCH (CC)}
  & \multicolumn{1}{c|}{GMM (CC)}
  & \multicolumn{1}{c|}{SpeClu (CC)}
  & \multicolumn{1}{c}{AutoSC (CC)} \\
\midrule
\textbf{ACC(default)} & 0.593 & 0.576 & 0.501 & 0.424 & 0.592 & 0.579 & 0.588  & 0.601 \\
\textbf{ACC(best)} & 0.636(\textcolor{red}{+0.043}) & 0.641(\textcolor{red}{+0.065}) & 0.631 (\textcolor{red}{+0.130}) & 0.570 (\textcolor{red}{+0.146}) & 0.619 (\textcolor{red}{+0.027}) & 0.626 (\textcolor{red}{+0.047}) & \cellcolor{orange!40} \textbf{0.688} (\textcolor{red}{+0.100}) & -\\
\midrule
\textbf{NMI(default)} & 0.336 & 0.311 & 0.178 & 0.028 & 0.330 & 0.315 & 0.318 & 0.321 \\
\textbf{NMI(best)} & 0.379(\textcolor{red}{+0.043}) & 0.373(\textcolor{red}{+0.062}) & 0.366 (\textcolor{red}{+0.188}) & 0.320 (\textcolor{red}{+0.292}) & 0.363 (\textcolor{red}{+0.033}) & 0.360 (\textcolor{red}{+0.045}) & \cellcolor{orange!40} \textbf{0.422} (\textcolor{red}{+0.104}) & -\\
\midrule
\textbf{ARI(default)} & 0.293 & 0.264 & 0.124 & 0.019 & 0.272 & 0.261 & 0.249 & 0.274\\
\textbf{ARI(best)}   & 0.344(\textcolor{red}{+0.051}) & 0.342(\textcolor{red}{+0.078}) & 0.323 (\textcolor{red}{+0.199}) & 0.256 (\textcolor{red}{+0.237}) & 0.316 (\textcolor{red}{+0.044}) & 0.318 (\textcolor{red}{+0.057}) & \cellcolor{orange!40} \textbf{0.380} (\textcolor{red}{+0.131}) & -\\
\midrule
\midrule
\textbf{Algorithms}  & \multicolumn{1}{c|}{SSC (CC)} & \multicolumn{1}{c|}{k-FSC (CC)} & \multicolumn{1}{c|}{k-PC (CC)} & \multicolumn{1}{c|}{MeanShift (CC)}  & \multicolumn{1}{c|}{S$^3$COMP-C (CC)} & \multicolumn{1}{c|}{LRR (CC)} & \multicolumn{1}{c|}{DEC (DC)} & \multicolumn{1}{c}{IDEC (DC)}\\
\midrule
\textbf{ACC(default)} & 0.518 & 0.496 & 0.445 & 0.398 & 0.517 & 0.463 & 0.560 & 0.550  \\
\textbf{ACC(best)} & 0.570(\textcolor{red}{+0.052}) & 0.579 (\textcolor{red}{+0.083}) & 0.466 (\textcolor{red}{+0.021}) & 0.485 (\textcolor{red}{+0.087})  & 0.549 (\textcolor{red}{+0.032}) & 0.529(\textcolor{red}{+0.066}) & 0.589(\textcolor{red}{+0.029}) & 0.603 (\textcolor{red}{+0.053}) \\
\midrule
\textbf{NMI(default)} & 0.200 & 0.200 & 0.121 & 0.105 & 0.187 & 0.144 & 0.290 & 0.251 \\
\textbf{NMI(best)} & 0.214(\textcolor{red}{+0.014}) & 0.249 (\textcolor{red}{+0.049}) & 0.136 (\textcolor{red}{+0.015}) & 0.221 (\textcolor{red}{+0.115}) & 0.210 (\textcolor{red}{+0.023}) & 0.200(\textcolor{red}{+0.056})& 0.316(\textcolor{red}{+0.026}) & 0.309 (\textcolor{red}{+0.058}) \\
\midrule
\textbf{ARI(default)} & 0.150 & 0.156 & 0.082 & 0.073 & 0.142& 0.108 & 0.248 & 0.210 \\
\textbf{ARI(best)} & 0.173(\textcolor{red}{+0.023}) & 0.218 (\textcolor{red}{+0.062}) & 0.101 (\textcolor{red}{+0.019}) & 0.172 (\textcolor{red}{+0.098}) & 0.172 (\textcolor{red}{+0.030}) &  0.165(\textcolor{red}{+0.057}) & 0.284(\textcolor{red}{+0.036}) & 0.278 (\textcolor{red}{+0.068}) \\
\midrule
\midrule
\textbf{Algorithms}  & \multicolumn{1}{c|}{DSCN (DC)} & \multicolumn{1}{c|}{PICA (DC)} & \multicolumn{1}{c|}{ConClu (DC)} & \multicolumn{1}{c|}{EDESC (DC)} & \multicolumn{1}{c|}{DMICC (DC)} & \multicolumn{1}{c|}{DIVC (DC)} & \multicolumn{1}{c|}{P$^2$OT (DC)} &  \multicolumn{1}{c}{LFSS (DC)}  \\
\midrule
\textbf{ACC(default)} & 0.550 & 0.540 & 0.519 & 0.557 & 0.543 & 0.541 & 0.546 & 0.529 \\
\textbf{ACC(best)}    & 0.600 (\textcolor{red}{+0.050}) & 0.599(\textcolor{red}{+0.059}) & 0.587 (\textcolor{red}{+0.068}) & \cellcolor{orange!40} \textbf{0.622} (\textcolor{red}{+0.065}) & 0.593 (\textcolor{red}{+0.050}) & 0.596(\textcolor{red}{+0.055}) & 0.589(\textcolor{red}{+0.043})& 0.579 (\textcolor{red}{+0.050}) \\
\midrule
\textbf{NMI(default)}  & 0.240 & 0.257 & 0.257 & 0.307 & 0.272 & 0.257 & 0.298 & 0.252 \\

\textbf{NMI(best)}    & 0.310 (\textcolor{red}{+0.070}) & 0.337(\textcolor{red}{+0.08}) & 0.321 (\textcolor{red}{+0.064}) & \cellcolor{orange!40}\textbf{0.367} (\textcolor{red}{+0.060}) & 0.317 (\textcolor{red}{+0.045}) & 0.339(\textcolor{red}{+0.082}) & 0.331(\textcolor{red}{+0.033}) & 0.305 (\textcolor{red}{+0.053}) \\
\midrule
\textbf{ARI(default)}  & 0.171 & 0.220 & 0.202 & 0.257 & 0.232 & 0.219 & 0.245 & 0.212 \\

\textbf{ARI(best)}   & 0.248 (\textcolor{red}{+0.077}) & 0.296(\textcolor{red}{+0.076}) & 0.283 (\textcolor{red}{+0.081}) & \cellcolor{orange!40}\textbf{0.333} (\textcolor{red}{+0.076}) & 0.287 (\textcolor{red}{+0.055}) & 0.296(\textcolor{red}{+0.077}) & 0.289(\textcolor{red}{+0.044}) & 0.275 (\textcolor{red}{+0.063}) \\

\bottomrule
\end{tabular}
}
\end{table*}

In this section, we compare the overall clustering performance (measured by ACC, NMI, and ARI) across 131 datasets. Table~\ref{tab-all-avg-p} reports the average performance over all datasets for both conventional clustering (CC) and deep clustering (DC) algorithms. For each algorithm, results under default hyperparameter settings and best-tuned hyperparameter configurations (HPCs) are provided to evaluate both robustness and performance ceilings, where the best HPC means selecting the best performance for each dataset. In addition, the statistical performance differences among all algorithms under the best HPC are illustrated in Appendix~\ref{fig-cd-diagram}. The detailed experimental settings are provided in Appendix~\ref{exp-settings}. Based on these results, we draw the following observations:

\begin{itemize}
    \item \textbf{Spectral clustering} (SpeClu) significantly \textbf{outperforms all other} algorithms, including conventional and deep clustering methods, in terms of average performance under the best hyperparameter configuration (HPC). Moreover, SpeClu is statistically indistinguishable (CD diagram in Appendix~\ref{fig-cd-diagram}) from a small subset of top-performing methods, as indicated by the non-overlapping CD intervals. This demonstrates that SpeClu not only attains strong absolute performance but also exhibits stable superiority across diverse datasets and evaluation criteria.

    \item All evaluated \textbf{deep clustering methods do not exhibit a significant advantage} over top-performing conventional algorithms in terms of average performance across the 131 datasets. We attribute this to two factors: (1) tabular data or feature embeddings lack explicit spatial structure and do not naturally support data augmentation strategies, which limits the transferability of deep models originally designed for image clustering; and (2) tabular features often directly capture semantic differences, such that simple similarity measures (e.g., Euclidean distance) remain highly effective, leaving limited room for neural representation learning to provide additional benefits.

    \item A notable observation from Table~\ref{tab-all-avg-p} is the substantial performance improvement from default to best configurations for both CC and DC methods.
    This phenomenon highlights the \textbf{high sensitivity of clustering algorithms to hyperparameter selection} and implies the importance of effective algorithm and HPC selection. Further analysis and exploration of model selection are provided in Section~\ref{analysis-ms}.

    \item The ACC, NMI, and ARI values are moderate even for top methods like SpeClu, reflecting the intrinsic difficulty of clustering diverse real-world tabular data. This highlights the challenge posed by the benchmark datasets and the ongoing need for clustering research.
\end{itemize}

The overall average results demonstrate that conventional clustering methods provide a higher performance ceiling than deep clustering methods across diverse tabular, text, and image datasets when hyperparameters are carefully tuned. However, this performance gain comes at the cost of model sensitivity. These findings suggest that future clustering research should focus on the development of effective automatic model selection and the design of deep clustering frameworks tailored for tabular data. Additional comparisons (statistical significance, performance distributions, computational cost) are in Appendix~\ref{appe-overall-analysis}.

\subsection{FM-based Clustering Analysis}
\label{analysis-fm}

In this section, we investigate clustering performance under foundation model (FM)-based settings. Guided by data modality and the current state of research, we design a series of experiments to systematically analyze clustering behavior across image, text, and tabular datasets.

\paragraph{Image datasets.}
For image clustering, each dataset is evaluated in both multiple pretrained feature embedding spaces and the original image data space. Specifically, we extract representations using three widely adopted pretrained models: ResNet~\cite{He2015} (ResNet18, ResNet50)~\footnote{https://github.com/KaimingHe/deep-residual-networks}, and CLIP (ViT-B/32)~\footnote{https://github.com/openai/CLIP}~\cite{radford2021CLIP}. We compare several top-performing baselines (identified in Section~\ref{analysis-1}) operating on these feature embeddings with state-of-the-art image-based deep clustering methods~\citep{dec, edesc, LFSS, li2025you}.
The corresponding results are summarized in Table~\ref{tab-image-performance-1}. First, utilizing pretrained representations leads to substantial performance improvements over operating directly on the original image data in the vast majority of cases, confirming the critical role of representation quality in image clustering. Second, conventional clustering algorithms such as KMeans and SpeClu achieve surprisingly strong performance when combined with high-quality pretrained embeddings, often rivaling or even surpassing specialized deep clustering methods. Third, state-of-the-art deep clustering approaches (e.g., LFSS~\cite{LFSS}, DCBoost~\cite{li2025you}) still demonstrate clear advantages on more challenging benchmarks such as CIFAR-10 and CIFAR-20, where learning dataset-specific structure remains essential.

Despite these impressive empirical gains, the results also raise important concerns regarding \emph{implicit data leakage}. Both ResNet and CLIP are pretrained on large-scale datasets (e.g., ImageNet-1K or web-scale image--text corpora), and benchmarks such as ImageNet-10 and ImageNet-Dogs are semantically overlapping subsets of these pretraining datasets. Consequently, clustering performance on these embeddings may partially benefit from latent class information already encoded during supervised or weakly supervised pretraining, although no labels are explicitly used during clustering.

\begin{table*}[h!]
\centering
\caption{Clustering comparison on image datasets with CLIP representations (Overall results are provided in Appendix~\ref{appe-fm}). The $*$ indicates the performance values reported in the original paper, which were obtained using the original image data. The best performances on each dataset are marked in \textcolor{orange}{\textbf{bold}}.} 
\label{tab-image-performance-1}
\resizebox{\textwidth}{!}{
\begin{tabular}{l|ccc|ccc|ccc|ccc|ccc|ccc}
\toprule
& \multicolumn{3}{c|}{\textbf{STL-10}} & \multicolumn{3}{c|}{\textbf{CIFAR-10}} & \multicolumn{3}{c|}{\textbf{CIFAR-20}} & \multicolumn{3}{c|}{\textbf{ImageNet-10 }} &\multicolumn{3}{c|}{\textbf{ImageNet-Dogs}} &\multicolumn{3}{c}{\textbf{Avg.}}\\
\cmidrule(lr){2-4}\cmidrule(lr){5-7}\cmidrule(lr){8-10}\cmidrule(lr){11-13}\cmidrule(lr){14-16} \cmidrule(lr){17-19}
{Method}
& NMI & ACC & ARI & NMI & ACC & ARI
& NMI & ACC & ARI & NMI & ACC & ARI
& NMI & ACC & ARI & NMI & ACC & ARI \\
\midrule
DEC$^*$ (ICML2016)  & 27.6 & 35.9 & 18.6 & 25.7 & 30.1 & 16.1 & 13.6 & 18.5 & 5.0  & 28.2 & 38.1 & 20.3 & 12.2 & 19.5 & 7.9 & 21.4 & 28.4 & 13.5\\
EDESC$^*$ (CVPR2022)& 68.7 & 74.5 & - & 46.4 & 62.7 & - & 37.0 & 38.5 & - & - & - & - & - & - & - & 50.7 & 58.5 & -\\
LFSS$^*$ (ICML2025)& 77.1 & 86.1 & 74.0 & 87.2 & 93.4 & 86.6 & 59.9 & 58.7 & 43.5 & 85.6 & 93.2 & 85.7 & 61.7 & 69.1 & 53.3 & 74.3 & 80.1 & 68.6\\
DCBoost$^*$ (NeurIPS2025)& \textbf{86.7} & \textbf{93.6} & \textbf{86.6} &
\cellcolor{orange!40}\textbf{91.1} & \cellcolor{orange!40}\textbf{96.0} & \cellcolor{orange!40}\textbf{91.6} &
\cellcolor{orange!40}\textbf{64.5} & \cellcolor{orange!40}\textbf{63.9} & \cellcolor{orange!40}\textbf{49.2} &
\textbf{92.7} & \textbf{97.1} & \textbf{93.7} &
\textbf{76.3} & \textbf{79.7} & \textbf{70.7} &
\cellcolor{orange!40}\textbf{82.3} & \cellcolor{orange!40}\textbf{86.1} & \cellcolor{orange!40}\textbf{78.4} \\
\midrule
DEC (CLIP) & 79.9 & 71.8 & 62.6 & 75.4 & 79.6 & 68.3 & 55.2 & 51.0 & 34.9 & 94.5 & 97.2 & 94.0 & 39.9 & 41.5 & 26.2 & 68.9 & 68.2 & 57.2\\
EDESC (CLIP) & 95.6 & 98.2 & 96.2 & \textbf{83.0} & 84.6 & 77.9 & \textbf{56.4} & \textbf{53.5} & \textbf{38.7} & \textbf{98.5} & \textbf{99.4} & \textbf{98.8} & 49.4 & 40.9 & 30.9 & \textbf{76.5} & 75.3 & 68.5\\
LFSS (CLIP) & 94.7 & 97.8 & 95.2 & 82.6 & \textbf{91.4} & \textbf{82.0} & 53.4 & 54.1 & 38.1 & 97.0 & 98.9 & 97.5 & 49.7 & 46.1 & 34.6 & 75.4 & \textbf{77.6} & \textbf{69.5}\\
KMeans (CLIP) & 95.1 & 98.0 & 95.7 & 78.7 & 86.6 & 70.6 & 52.7 & 52.2 & 34.3 & 97.5 & 99.0 & 97.9 & \textbf{50.9} & \textbf{51.4} & \textbf{35.7} & 74.9 & 77.4 & 66.8\\
SpeClu (CLIP) &
\cellcolor{orange!40}\textbf{96.3} & \cellcolor{orange!40}\textbf{98.5} & \cellcolor{orange!40}\textbf{96.8} &
79.0 & 85.4 & 67.3 &
47.3 & 46.4 & 28.8 &
98.2 & 99.4 & 98.6 &
48.4 & 51.6 & 35.1 &
73.8 & 76.2 & 65.3\\
\bottomrule
\end{tabular}}
\end{table*}

\begin{table*}[h!]
\centering
\caption{Clustering comparison on text datasets with embeddings derived from different foundation models. The best performances are marked in \textcolor{orange!40}{bold}.}
\label{tab-text-performance}
\resizebox{\textwidth}{!}{
\begin{tabular}{l|ccc|ccc|ccc|ccc|ccc|ccc}
\toprule
& \multicolumn{3}{c|}{\textbf{20Newsgroups}} & \multicolumn{3}{c|}{\textbf{Enron}} & \multicolumn{3}{c|}{\textbf{IMDB}} & \multicolumn{3}{c|}{\textbf{Reuters21578}} &\multicolumn{3}{c|}{\textbf{WOS}} &\multicolumn{3}{c}{\textbf{Avg.}}\\
\cmidrule(lr){2-4}\cmidrule(lr){5-7}\cmidrule(lr){8-10}\cmidrule(lr){11-13}\cmidrule(lr){14-16}\cmidrule(lr){17-19}
{Method}
& NMI & ACC & ARI & NMI & ACC & ARI
& NMI & ACC & ARI & NMI & ACC & ARI
& NMI & ACC & ARI & NMI & ACC & ARI\\
\midrule
DEC (BERT) &  0.091 & 0.123 & 0.030 & 0.055 & 0.620 & 0.058 & \textbf{0.035} & \textbf{0.608} & \textbf{0.049} & 0.176 & 0.577 & 0.177 & 0.133 & 0.309 & 0.106 & 0.098 & 0.447 & 0.084 \\
EDESC (BERT) & 0.205 & 0.187 & 0.101 & 0.067 & 0.643 & 0.090 & 0.009 & 0.558 & 0.013 & \textbf{0.339} & \textbf{0.663} & \textbf{0.333} & 0.203 & 0.374 & 0.197 & 0.165 & 0.485 & 0.147\\
KMeans (BERT) & 0.149 & 0.150 & 0.049 & 0.032 & 0.588 & 0.031 & 0.004 & 0.538 & 0.005 & 0.157 & 0.534 & 0.172 & 0.201 & 0.310 & 0.120 & 0.109 & 0.424 & 0.075\\
SpeClu (BERT) & \cellcolor{orange!40}\textbf{0.665} & \textbf{0.584} & \textbf{0.496} & \textbf{0.122} & \textbf{0.677} & \textbf{0.126} & 0.020 & 0.584 & 0.028 & 0.144 & 0.497 & 0.121 & \textbf{0.348} & \textbf{0.495} & \textbf{0.277} & \textbf{0.260} & \textbf{0.567} & \textbf{0.210}\\
\midrule
DEC (Llama3) & 0.488 & 0.415 & 0.267 & 0.034 & 0.572 & 0.028 & 0.075 & 0.655 & 0.100 & 0.541 & 0.797 & 0.504 & 0.423 & 0.542 & 0.324 & 0.312 & 0.596 & 0.245\\
EDESC (Llama3) &  0.620 & 0.595 & 0.477 & \textbf{0.356} & \textbf{0.802} & \textbf{0.424} & 0.078 & 0.662 & 0.106 & \textbf{0.738} & \textbf{0.920} & \textbf{0.782} & 0.463 & 0.612 & 0.372 & 0.451 & 0.718 & 0.432\\
KMeans (Llama3) & 0.623 & 0.627 & 0.454 & 0.073 & 0.627 & 0.065 & 0.202 & 0.755 & 0.260 & 0.626 & 0.875 & 0.665 & 0.461 & 0.604 & 0.365 & 0.397 & 0.698 & 0.362\\
SpeClu (Llama3) & \textbf{0.649} & \cellcolor{orange!40}\textbf{0.670} & \cellcolor{orange!40}\textbf{0.518} & 0.287 & 0.802 & 0.365 & \textbf{0.436} & \textbf{0.861} & \textbf{0.523} & 0.508 & 0.709 & 0.444 & \textbf{0.469} & \textbf{0.624} & \textbf{0.380} & \textbf{0.470} & \textbf{0.733} & \textbf{0.446}\\
\midrule
DEC (OpenAI) &  0.549 & 0.521 & 0.367 & 0.773 & 0.958 & 0.842 & 0.249 & 0.775 & 0.304 & 0.617 & 0.829 & 0.576 & 0.441 & 0.546 & 0.325 & 0.526 & 0.726 & 0.483\\
EDESC (OpenAI) & 0.617 & 0.597 & \textbf{0.485} & \cellcolor{orange!40}\textbf{0.876} & \cellcolor{orange!40}\textbf{0.981} & \cellcolor{orange!40}\textbf{0.927} & 0.333 & 0.825 & 0.424 & \cellcolor{orange!40}\textbf{0.821} & \cellcolor{orange!40}\textbf{0.953} & \cellcolor{orange!40}\textbf{0.871} & \cellcolor{orange!40}\textbf{0.474} & 0.632 & 0.383 & \cellcolor{orange!40}\textbf{0.624} & 0.798 & \cellcolor{orange!40}\textbf{0.618}\\
KMeans (OpenAI) & 0.621 & 0.635 & 0.446 & 0.842 & 0.971 & 0.890 & 0.362 & 0.834 & 0.446 & 0.742 & 0.923 & 0.788 & 0.459 & 0.588 & 0.354 & 0.605 & 0.790 & 0.585\\
SpeClu (OpenAI) & \textbf{0.622} & \textbf{0.646} & 0.454 & 0.788 & 0.957 & 0.838 & \cellcolor{orange!40}\textbf{0.482} & \cellcolor{orange!40}\textbf{0.882} & \cellcolor{orange!40}\textbf{0.584} & 0.667 & 0.889 & 0.695 & 0.456 & \cellcolor{orange!40}\textbf{0.644} & \cellcolor{orange!40}\textbf{0.389} & 0.603 & \cellcolor{orange!40}\textbf{0.804} & 0.592\\

\bottomrule
\end{tabular}
}
\end{table*}

\begin{table*}[h!]
\centering
\caption{Clustering comparison on tabular datasets between non-LLM methods and LLMs based on prompt learning. The best performances are marked in \textcolor{orange!40}{bold}.}
\label{tab-tabular-performance}

\resizebox{\textwidth}{!}{
\begin{tabular}{l|ccc|ccc|ccc|ccc|ccc|ccc|ccc}
\toprule
& \multicolumn{3}{c|}{\textbf{echocardiogram}} & \multicolumn{3}{c|}{\textbf{world12d}} & \multicolumn{3}{c|}{\textbf{hepatitis}} & \multicolumn{3}{c|}{\textbf{zoo}} &\multicolumn{3}{c|}{\textbf{spectf$\_$heart}} &\multicolumn{3}{c|}{\textbf{image$\_$segmentation}} & \multicolumn{3}{c}{\textbf{Avg.}}\\
\cmidrule(lr){2-4}\cmidrule(lr){5-7}\cmidrule(lr){8-10}\cmidrule(lr){11-13}\cmidrule(lr){14-16}\cmidrule(lr){17-19}\cmidrule(lr){20-22}
{Method}
& NMI & ACC & ARI & NMI & ACC & ARI
& NMI & ACC & ARI & NMI & ACC & ARI
& NMI & ACC & ARI  & NMI & ACC & ARI & NMI & ACC & ARI\\
\midrule
DEC &  0.562 & 0.907 & 0.651 & 0.717 & 0.777 & 0.694 & 0.160 & 0.812 & 0.292 & 0.756 & 0.778 & 0.688 & 0.065 & 0.608 & 0.065 & 0.611 & 0.550 & 0.419 & 0.479 & 0.739 & 0.468 \\
EDESC  & 0.578 & 0.890 & 0.604 & 0.698 & 0.782 & 0.678 & 0.173 & 0.737 & 0.201 & 0.827 & \cellcolor{orange!40}\textbf{0.815} & \cellcolor{orange!40}\textbf{0.832} & 0.127 & 0.695 & 0.149 & 0.576 & 0.552 & 0.427 & 0.497 & 0.745 & 0.482\\
KMeans & 0.505 & 0.868 & 0.536 & 0.776 & 0.815 & 0.754 & 0.177 & 0.775 & 0.254 & \cellcolor{orange!40}\textbf{0.843} & 0.808 & 0.742 & 0.221 & 0.741 & 0.228 & 0.616 & 0.666 & 0.493 & 0.523 & 0.779 & 0.501\\
SpeClu  & \textbf{0.639} & \textbf{0.918} & \textbf{0.690} & \cellcolor{orange!40}\textbf{0.842} & \cellcolor{orange!40}\textbf{0.920} & \cellcolor{orange!40}\textbf{0.850} & \cellcolor{orange!40}\textbf{0.241} & \cellcolor{orange!40}\textbf{0.837} & \cellcolor{orange!40}\textbf{0.382} & 0.791 & 0.792 & 0.705 & \textbf{0.389} & \textbf{0.762} & \textbf{0.350} & \textbf{0.689} & \textbf{0.744} & \textbf{0.581} & \cellcolor{orange!40}\textbf{0.599} & \cellcolor{orange!40}\textbf{0.829} & \cellcolor{orange!40}\textbf{0.593}\\
\midrule
Llama4 & 0.001 & 0.590 & -0.023 & \textbf{0.478} & \textbf{0.500} & 0.218 & \textbf{0.191} & \textbf{0.712} & \textbf{0.166} & 0.216 & 0.376 & 0.089 & \cellcolor{orange!40}\textbf{0.493} & \cellcolor{orange!40}\textbf{0.887} & \cellcolor{orange!40}\textbf{0.595} & \cellcolor{orange!40}\textbf{1.000} & \cellcolor{orange!40}\textbf{1.000} & \cellcolor{orange!40}\textbf{1.000} & \textbf{0.397} & \textbf{0.678} & \textbf{0.341}\\
DeepSeek-V3.2 & 0.126 & 0.754 & 0.224 & 0.456 & 0.486 & \textbf{0.250} & 0.127 & 0.525 & -0.072 & \textbf{0.328} & \textbf{0.574} & \textbf{0.192} & 0.001 & 0.525 & -0.010  & 0.921 & 0.790 & 0.803 & 0.327 & 0.609 & 0.231\\
GPT5-mini & \cellcolor{orange!40}\textbf{0.686} & \cellcolor{orange!40}\textbf{0.934} & \cellcolor{orange!40}\textbf{0.747} & 0.003 & 0.220 & -0.022 & 0.009 & 0.537 & -0.001 & 0.107 & 0.257 & -0.005 & 0.000 & 0.500 & -0.012 & 0.002 & 0.166 & -0.027 & 0.135 & 0.436 & 0.113\\

\bottomrule
\end{tabular}
}
\end{table*}

\paragraph{Text datasets.}
For text clustering, we adopt three pretrained foundation models, BERT (bert-large-uncased)~\footnote{https://github.com/google-research/bert}~\cite{devlin2018bert}, Llama3~\cite{llama3modelcard}, and OpenAI’s text-embedding-3-large~\footnote{https://platform.openai.com/docs/guides/embeddings}, to extract text embeddings. The corresponding results are reported in Table~\ref{tab-text-performance}. Overall, using more advanced foundation models for embedding extraction consistently leads to larger performance gains across all metrics. SpeClu and EDESC substantially benefit from enhanced semantic representations, achieving superior NMI, ACC, and ARI on most datasets.

\paragraph{Tabular datasets.}
For tabular data, existing foundation models such as TabPFN~\cite{hollmann2023tabpfn,hollmann2025tabpfn} are primarily designed for supervised learning and are not directly applicable to unsupervised clustering tasks. Moreover, many of the collected tabular datasets lack feature names, which hinders the construction of meaningful semantic prompts to convert tabular inputs into natural language descriptions suitable for LLM processing. As a result, we resort to in-context learning with LLMs to perform clustering on tabular datasets. The prompt template is provided in Appendix~\ref{appe-fm}. Due to context length constraints of LLMs, we evaluate them on six small-scale tabular datasets and the results are reported in Table~\ref{tab-tabular-performance}. LLM-based methods outperform traditional clustering approaches on three datasets, while SpeClu attains the best average performance overall. These findings highlight, on the one hand, the promising potential of LLMs for fully unsupervised reasoning tasks, and on the other hand, the continued necessity of non-LLM clustering research for tabular data.

Based on the empirical results and analyses above, we draw the following conclusions:
\begin{itemize}
    \item From a practical perspective, for image and text clustering tasks, combining pretrained embeddings with conventional clustering algorithms (e.g., KMeans, SpeClu) offers an effective and computationally efficient solution and should be considered as a strong baseline.

    \item From a methodological perspective, existing tabular foundation models exhibit notable limitations, particularly for unsupervised learning scenarios.
    \item From an academic perspective, clustering remains challenging, even in the era of dominant foundation models. Notably, potential data leakage from pretrained models must be carefully accounted for to ensure fair and meaningful comparisons.

\end{itemize}

\subsection{Similarity Analysis of Algorithms and Datasets}
\label{analysis-3}

To analyze performance similarities among clustering algorithms across diverse datasets and evaluation metrics, as well as similarities among datasets across different algorithms and metrics, we construct \emph{performance vectors} at both the algorithm and dataset levels. Specifically, each algorithm is represented by a vector $\mathbf{p} \in \mathbb{R}^{N}$, where each entry corresponds to its performance (e.g., ACC) on a particular dataset and $N$ denotes the total number of datasets in our benchmark. Conversely, each dataset is characterized by a vector $\mathbf{p} \in \mathbb{R}^{M}$ that captures the performance of all evaluated clustering algorithms on a given dataset, with $M$ denoting the total number of methods considered. Aggregating these vectors yields three performance matrices, namely $\mathbf{P}_{\text{acc}}, \mathbf{P}_{\text{nmi}}, \mathbf{P}_{\text{ari}} \in \mathbb{R}^{N \times M}$, corresponding to ACC, NMI, and ARI, respectively. To obtain a holistic performance representation that jointly reflects all evaluation criteria, we concatenate these matrices to form the overall performance matrix $\mathbf{P}_\text{all} = [\mathbf{P}_{\text{acc}}^{\top}; \mathbf{P}_{\text{nmi}}^{\top};\mathbf{P}_{\text{ari}}^{\top}] \in \mathbb{R}^{M \times 3N}$, where $[\cdot;\cdot;\cdot]$ denotes the row-wise concatenation of three matrices and each row $[\mathbf{P}_\text{all}]_{i,:}$ serves as a comprehensive performance vector for the $i$-th algorithm. To visualize and interpret similarities among these high-dimensional performance vectors, we apply t-SNE~\cite{maaten2008visualizing} to project them into a two-dimensional space. The analysis and visualization among datasets are provided in Appendix~\ref{appe-simi-data}.

\begin{figure}[!ht]
    \centering
   \includegraphics[width=0.6\linewidth]{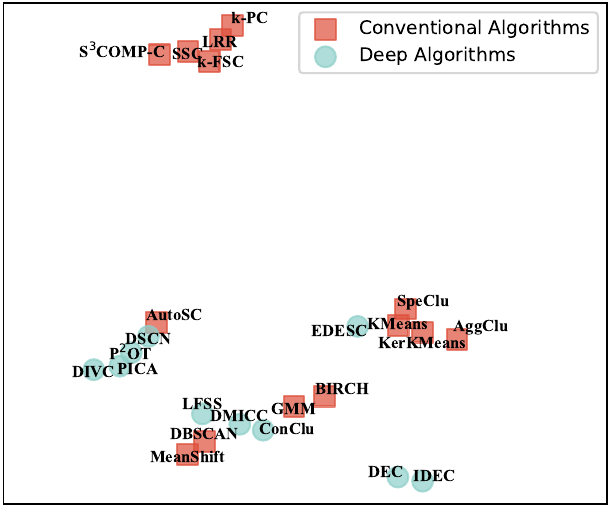}
     \caption{The t-SNE visualization results on algorithm performance vectors. }
    \label{fig-tsne-methods}
\end{figure}

\paragraph{Among Algorithms.}
In Figure~\ref{fig-tsne-methods}, algorithms with closely related methodological designs appear in close proximity in the embedding space. For example, subspace approaches (LRR, SSC, k-FSC, and $\mathrm{S}^3$COMP-C) form a tight group, indicating shared assumptions and optimization principles. IDEC appears near DEC, consistent with IDEC extending DEC, and DIVC is close to PICA, matching their architectural and objective-level similarity.
These observed performance clusters provide both methodological and practical insights: they enable a principled, performance-driven categorization of clustering algorithms beyond design-based taxonomies, and they imply that testing a few representative methods per cluster can approximate the broader performance landscape on new datasets, reducing experimental cost.
Overall, Figure~\ref{fig-tsne-methods} demonstrates that clustering algorithms exhibit strong and interpretable performance regularities, and that performance-vector-based visualization serves as a powerful tool for understanding algorithmic relationships.

\begin{figure}
    \centering
    \includegraphics[width=\linewidth]{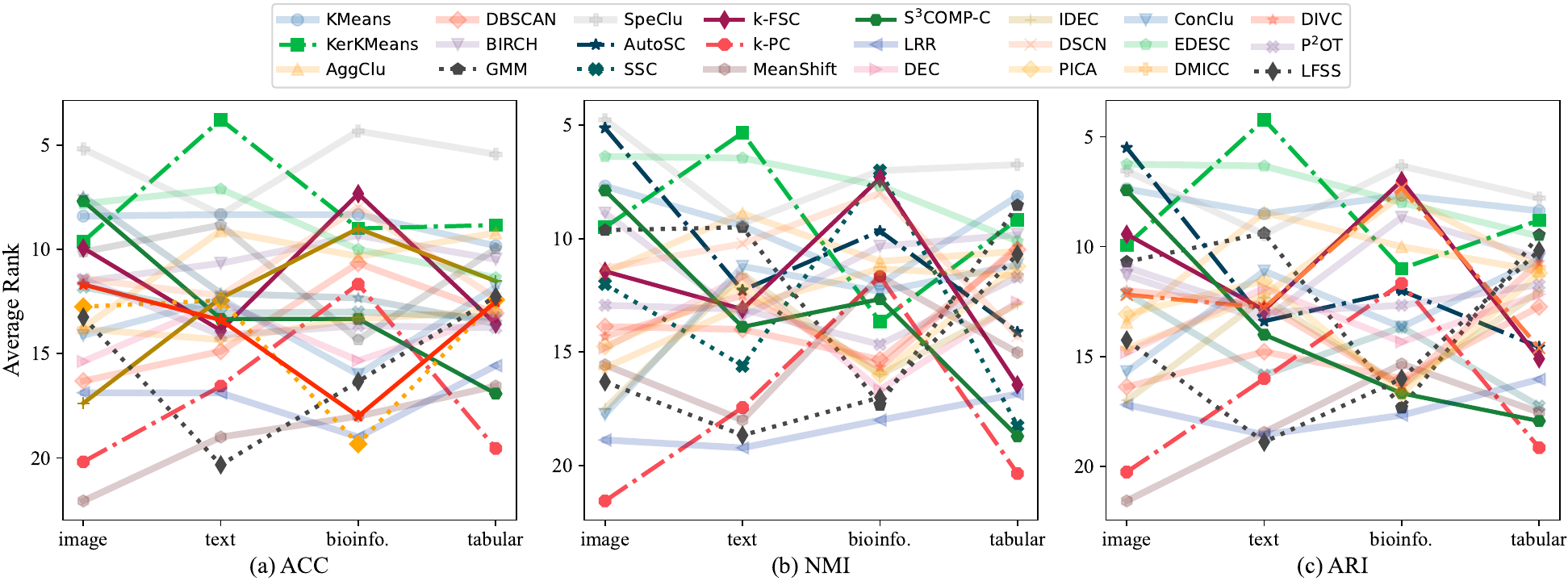}
    \caption{Algorithm preference analysis grouped by data types under three evaluation metrics. The top eight methods exhibiting the largest performance fluctuations are highlighted.}
    \label{fig-preference-type}
\end{figure}

\subsection{Algorithm Preference Analysis}
\label{analysis-2}

Beyond the overall performance analysis, we further investigate algorithm preference by examining how algorithm rankings vary across datasets with different characteristics. Specifically, we group datasets from three complementary perspectives: (i) data type (image, text, tabular, and bioinformatics). We treat bioinformatics datasets as a separate data category because they typically exhibit extremely high feature dimensionality, often with the number of features exceeding the number of samples. This characteristic distinguishes them substantially from other tabular datasets and leads to markedly different clustering behavior; (ii) feature dimensionality, categorized as low ($m \leq 100$), middle ($100 < m \leq 500$), and high ($m > 500$); and (iii) the degree of cluster size imbalance. 
Figure~\ref{fig-preference-type} visualizes average performance ranks under data type categorization. Other visualizations under different data groupings are provided in Appendix~\ref{appe-preference}. These results demonstrate that algorithm performance rankings are highly context-dependent. Feature dimensionality, data modality, and cluster imbalance each induce distinct preference patterns, and algorithms with strong overall performance may still be suboptimal under specific dataset conditions (e.g., SpeClu). These findings underscore the importance of conditional algorithm selection rather than relying solely on global rankings. From a practical standpoint, the analysis provides actionable guidance for selecting clustering algorithms tailored to dataset characteristics, while from a research perspective, it highlights open challenges in designing methods that are simultaneously robust across diverse data regimes.

\subsection{Low-rank Analysis on Performance Matrices}
\label{analysis-4}
To analyze the low-rank structure of the performance matrices, we perform singular value decomposition (SVD) on the performance matrices (ACC, NMI, ARI) where each performance matrix consists of 131 rows (datasets) and 273 columns (all clustering algorithms under different HPCs).

Let $\mathbf{\hat{P}}_{acc}, \mathbf{\hat{P}}_{nmi}, \mathbf{\hat{P}}_{ari}\in\mathbb{R}^{N\times H}$ be the corresponding performance matrices where $\sigma_i$ denotes the $i$-th singular value, where $\sigma_1 \geq \sigma_2 \geq \cdots \geq \sigma_n > 0$, $N=131$ denotes the number of datasets, $H=273$ denotes the number of all clustering algorithms under different hyperparameter configurations. Figure~\ref{fig-singular-ccr} (a) presents the cumulative contribution ratio of the singular values of $\mathbf{\hat{P}}_{acc}$, where the cumulative contribution ratio ($ccr$) is defined by $ccr(j) = (\sum_{i=1}^j \sigma_i) / (\sum_{i=1}^n \sigma_i)$. 
As evidenced by Figure~\ref{fig-singular-ccr} (a), the cumulative contribution ratio of the first sixty (60/131) singular values ($ccr(60)$) exceeds $90\%$. This demonstrates that the performance matrix $\mathbf{\hat{P}}_{acc}$ possesses a low-rank structure. The clustering performance of new datasets or methods can be reliably predicted from a subset of their measurements, enabling an efficient approximation for overall performance evaluation in practical applications.

To further verify the low-rank property of $\mathbf{\hat{P}}_{acc}$ and the effectiveness of performance prediction, we construct matrix completion tasks under the MCAR (missing completely at random) mechanism with missing rates $ \textbf{mr} \in \{0.5, 0.6, 0.7, 0.8, 0.9\}$. We use matrix factorization and non-convex optimization techniques~\citep{candes2012exact, chi2018low, fan2019factor} to recover the missing entries of the performance matrix $\mathbf{\hat{P}}_{acc}$. The recovery results are provided in Table~\ref{acc-matrix-completion}, indicating that rapid and reliable performance prediction is possible based on the performance matrix $\mathbf{\hat{P}}_{acc}$. The low-rank analysis of the three performance matrices and the detailed matrix completion process are illustrated in Appendix~\ref{appe-low-rank}.

\begin{table}[!ht]
    \centering
    \caption{Recovery performance (MAPE) on the performance matrix $\mathbf{P}_{acc}$ in the setting of MCAR.}
    \label{acc-matrix-completion}
    
    \begin{tabular}{|c|c|c|c|c|c|}
    \hline
         \textbf{mr} & 0.5 & 0.6 & 0.7 & 0.8 & 0.9 \\
    \hline
         \multirow{2}{*}{\textbf{MAPE}} & 0.1191 &  0.1326 & 0.1500 &  0.1750 & 0.2273 \\
        & (0.0090) & (0.0013) & (0.0021) & (0.0030)	& (0.0063) \\
    \hline
    \end{tabular}
\end{table}

\subsection{Model Selection Based on Performance Matrices}
\label{analysis-ms}

The results and analysis in Section~\ref{analysis-1} demonstrate a substantial performance gap among different hyperparameter configurations (HPCs) for both conventional and deep clustering methods. Moreover, as shown in Section~\ref{analysis-4}, the best-performing algorithms or HPCs for different datasets vary due to differences in their structure and meta-features. These two observations highlight the pronounced sensitivity of clustering algorithms to hyperparameter selection and motivate the need for model selection strategies. To this end, we construct a set of meta features $\mathbf{z}$ for each dataset and aim to learn an effective mapping $f$ from the meta-feature space $\mathbf{Z} = \{\mathbf{z}_1, \mathbf{z}_2, \ldots, \mathbf{z}_N\}$ to the clustering performance space (ACC, NMI and ARI). To learn the mapping $f$, we experiment with three regression models: XGBoost~\cite{chen2016xgboost}, LightGBM~\cite{ke2017lightgbm}, and Random Forest~\cite{breiman2001random}. Detailed descriptions of the meta-feature construction and the training procedures are provided in Appendix~\ref{appe-ms}.

Table~\ref{tab-model-selection} reports the average results of 5-fold cross-validation under different model selection strategies. In particular, \textcolor{red}{\textbf{Empirical Upper Bound (EUB)}} represents the optimal average performance achievable in our hyperparameter configuration space. \emph{KMeans} denotes a heuristic strategy that selects the hyperparameter configuration with the best historical average performance, with analogous procedures applied to the other baseline methods. As shown in Table~\ref{tab-model-selection}, all three regression-based selectors consistently achieve performance that is comparable to, and in some cases surpasses, the clustering baselines across all evaluation metrics. This result suggests that it is feasible to approximate the relationship between dataset characteristics and clustering performance through appropriately designed meta features. This verification experiment indicates that performance-driven model selection is a promising direction for automated clustering. In particular, constructing more informative meta features and adopting more expressive regression models may further improve the accuracy of $f$, narrowing the gap toward the empirical upper bound.

\begin{table}[!ht]
    \centering
    \caption{The verification experiments (5-fold cross-validation) of model selection based on performance matrices.}
    \label{tab-model-selection}
    \begin{tabular}{c|c|c|c}
    \toprule
    Methods & ACC & NMI & ARI \\
    \midrule
    \textcolor{red}{\textbf{EUB}} & \textcolor{red}{\textbf{74.65}} & \textcolor{red}{\textbf{50.86}} & \textcolor{red}{\textbf{48.65}} \\
    KMeans~\cite{mcqueen1967some} & 58.57 & 34.51 & 28.11 \\
    DEC~\cite{dec} & 56.07 & 30.13 & 25.60 \\
    DSCN~\cite{dscn} &  55.73 & 26.09 & 19.65 \\
    EDESC~\cite{edesc} & 57.81 & 33.44 & 28.54 \\
    DIVC~\cite{DIVC} & 55.36 & 29.87 & 24.83 \\
    P$^2$OT~\cite{zhang2024p} & 55.29 & 30.62 & 25.73 \\
    LFSS~\cite{LFSS} & 54.02 & 27.74 & 23.57 \\
    \midrule
    MS (XGBoost~\cite{chen2016xgboost}) & 58.31 & 33.47 & 26.70\\
    MS (LightGBM~\cite{ke2017lightgbm}) & 58.38 & 33.98 & 27.96\\
    MS (Random Forest~\cite{breiman2001random}) & \textbf{59.67} & \textbf{34.47} & \textbf{28.76} \\
    \bottomrule
    \end{tabular}
\end{table}

\section{Conclusions}

This paper introduced CLUBench, a comprehensive and challenging clustering benchmark that evaluates conventional, deep, and foundation model-based methods across three data modalities. Extensive experiments showed that deep clustering methods offer no significant advantage over top-performing conventional algorithms in average performance. Comparisons revealed that combining pretrained embeddings with conventional algorithms (e.g., KMeans, SpeClu) provides an effective and efficient solution for image and text clustering. Further analyses—including algorithm preference, performance-based similarity, low-rank structure, and model selection—uncovered valuable findings and suggested promising directions for future research.

\section*{Acknowledgements}
This work was partially supported by the National Natural Science Foundation of China under Grant No.62376236, General Program of Natural Science Foundation of Guangdong Province Grant No.2024A1515011771, and the Shenzhen Stability Science Program 2023.

\bibliographystyle{unsrtnat}
\bibliography{references}

\appendix

\section{Related Work}

\subsection{Clustering algorithms}
\label{app-CA}
\paragraph{\textbf{Conventional Clustering (CC) Algorithms.}}
Conventional clustering methods typically operate on original data features through distance calculation, similarity measurement, or density estimation. They can be categorized into $6$ groups: partition-based, hierarchical, density-based, 
model-based, graph-based and subspace-based. Representative examples include K-means~\citep{mcqueen1967some} for partitioning, BIRCH~\citep{zhang1996birch} for hierarchical clustering, DBSCAN~\citep{ester1996density} and OPTICS~\citep{ankerst1999optics} for density-based clustering, 
Gaussian Mixture Models (GMMs)~\citep{rasmussen1999infinite} for model-based clustering and spectral clustering~\citep{von2007tutorial} for graph-based clustering, and SSC~\citep{elhamifar2013sparse}, S$^3$COMP-C~\citep{chen2020stochastic} and  k-FSC~\citep{fan2021large} for subspace clustering. Besides these basic algorithms, there have been many extensions, such as kernel K-means \citep{liu2022simplemkkm}, low-rank representation \citep{liu2012robust}, multi-view subspace clustering \citep{Kang_Zhou_Zhao_Shao_Han_Xu_2020}, federated spectral clustering \citep{qiao2023federated}, etc.

\paragraph{\textbf{Deep Clustering (DC) Algorithms.}}In recent years, deep clustering~\citep{dec,dcgmm,idec,dscn,caron2018deep,han2019learning,asano2019self,pica,edesc,DIVC,zhang2024p,LFSS,VDE,JULE,DAC,DCC,DSEC,DDCA,niu2022spice,nie2023fast,do2021clustering,peng2025provable,guo2022hcsc,qian2023stable,li2023image,liu2024interactive,shen2021you,zhang2024p, zhu2025hierarchical, li2025you} has emerged to address the challenges of large-scale and complex structured datasets. Autoencoder-based methods~\citep{dec,idec,VDE} use reconstruction and clustering losses to learn meaningful embeddings for clustering. For instance, a seminal approach is deep embedded clustering (DEC)~\citep{dec}, which jointly learns feature representations and cluster assignments. Deep subspace clustering methods~\citep{dscn,edesc} learn pairwise affinities between pairs of data points and then use methods like spectral clustering for clustering. For instance, DSCN \citep{dscn} combines an autoencoder and a self-expressive layer to learn the pairwise affinity between each pair of data points. Contrastive learning-based methods~\citep{cc,DMICC,shen2021you,guo2022hcsc,liu2024interactive,shen2021you} usually use data augmentation techniques to create positive pairs and negative pairs, whose similarity is maximized or minimized by a contrastive loss. For instance, contrastive clustering (ConClu)~\citep{cc} uses instance-level and cluster-level contrastive learning to learn the predicted labels of each data point, while HCSC~\citep{guo2022hcsc} uses prototypes to learn hierarchical semantic structures in data and select more representative positive and negative pairs to learn better data representations. Information theory-based methods~\citep{pica,DIVC,do2021clustering,DCC} use neural networks to minimize cross entropy loss, negative entropy loss, mutual information loss, or other losses related to information theory. For instance, PICA~\citep{pica} minimizes the cross-entropy loss between the ground truth class and its prediction of training data, while negative entropy loss is used to avoid trivial solutions. CRLC~\citep{do2021clustering} maximizes the mutual information of data from different augmentations and combines clustering with contrastive learning to discriminate objects in the same cluster. Self-supervision-based methods \citep{LFSS,JULE,DAC,DCC,DSEC,DDCA,niu2022spice,qian2023stable,JULE} generate pseudo-labels for each data point, which are optimized together with the neural network during training. JULE~\citep{JULE} divides the deep clustering problem into conventional clustering and supervised representation learning, which then uses a recurrent framework to optimize them in the forward pass and backward pass, respectively. SPICE~\citep{niu2022spice} designs semantic-aware pseudo-labeling methods and combines contrastive learning to improve clustering performance. Subsequent works explore diverse directions: P2OT~\citep{zhang2024p} uses optimal transport to deal with the pseudo-labeling problem, which is effective in imbalanced clustering problems. Several other works like~\citep{li2023image,peng2025provable} use textual descriptions of image data as supervision signals to guide clustering. These methods demonstrate the rapid evolution of deep clustering, yet systematic evaluation remains limited. 

\subsection{Existing Clustering Benchmarks and Reviews with Experimental Evaluation}
\label{app-bench-related}
As discussed above, numerous conventional and deep clustering algorithms have been developed over the past decades. A systematic evaluation of these methods on diverse real-world datasets, together with an easy-to-use implementation toolbox, is therefore essential for advancing the field. Prior studies~\citep{jain1999data, xu2005survey, berkhin2006survey, omran2007overview, von2010clustering, murtagh2012algorithms, xu2015comprehensive, min2018survey, aljalbout2018clustering, nutakki2018introduction, javed2020benchmark, liu2022survey, yin2024rapid, ren2024deep, zhou2024comprehensive, wei2024overview, wehrli2023german, o2025openclustered} have reviewed clustering algorithms from various perspectives.
For example, \citet{jain1999data} reviews partition-based and hierarchical methods, discussing their theoretical foundations and applications. \citet{xu2005survey} evaluates both conventional and neural-network–based methods on real datasets and discusses applications in domains such as the traveling salesman problem and bioinformatics. \citet{murtagh2012algorithms} focuses on hierarchical clustering and provides implementations across multiple software environments. \citet{javed2020benchmark} benchmarks conventional methods on 112 diverse time-series datasets, while \citet{thrun2020clustering} introduces the Fundamental Clustering Problems Suite (FCPS), a curated dataset collection designed to expose algorithmic strengths and weaknesses under visually interpretable conditions. \citet{shand2021hawks} proposes HAWKS, an evolutionary framework for generating synthetic benchmark datasets that systematically span diverse clustering properties. 
More recent work emphasizes deep clustering. \citet{leiber2023benchmarking} focuses on deep clustering for image data and introduces Clustpy, a unified benchmarking framework for fair comparison. \citet{zhou2024comprehensive} decomposes deep clustering methods into representation learning and clustering components and proposes a taxonomy based on their interaction. \citet{wei2024overview} provides a comprehensive overview of deep clustering for graph and image data, analyzing architectures, loss formulations, and methodological categories, with evaluations on several datasets. Collectively, these studies offer valuable insights into the development and application of clustering techniques. However, despite their contributions, several limitations remain: evaluations are typically conducted on a limited set of algorithms and datasets; joint comparisons across traditional, deep learning, and foundation model–based methods are largely missing; and no unified, easy-to-use toolbox is available to support reproducible benchmarking. 
As summarized in Table~\ref{tab-benchmark-com}, compared with existing studies, CLUBench provides a substantially larger experimental scale, more comprehensive and in-depth analyses, and an easy-to-use toolbox.

\section{Experimental Coverage and Settings}
\label{exp-settings}

The experimental coverage of CLUBench is summarized in Table~\ref{exp-converage}.

\begin{table}[!ht]
    \centering
    \caption{The experimental coverage of CLUBench.}
    \label{exp-converage}
    \resizebox{0.7\textwidth}{!}{
    \begin{tabular}{|c|c|c|c|c|c|}
    \hline
    \# Datasets & \# Algorithms & \# HPCs & \# Exp. Repeat & \# Total Exp. & Metrics \\
    \hline
     131 & 24 & 273 & 5 & 178,815(131 $\times$ 273 $\times$ 5) & ACC, NMI, ARI \\
    \hline
    \end{tabular}
    }
\end{table}

\paragraph{Settings.} In CLUBench, we focus only on hard clustering and use the same settings as most existing deep clustering methods~\cite{dec, idec, edesc, dscn, pica, DIVC,cc, zhang2024p, DMICC,LFSS}, in which the ground-truth number of clusters $K$ is provided to the methods. DBSCAN, a standard implementation in \texttt{scikit-learn}~\footnote{https://scikit-learn.org/stable/}, does not require $K$ as an input hyperparameter. Consequently, we tuned its hyperparameter configurations over a broader range, resulting in 90 distinct configurations. This extensive tuning is intended to provide a relatively fair comparison with methods that rely on a predefined number of clusters $K$.

\paragraph{Evaluation.} 
Following previous studies~\cite{dec, DIVC, zhang2024p, LFSS, li2025you}, we evaluate all clustering methods in CLUBench using ACC (Clustering Accuracy), NMI (Normalized Mutual Information), and ARI (Adjusted Rand Index). Internal evaluation metrics, such as Silhouette Coefficient (SC), Davies-Bouldin Index (DBI), and Calinski-Harabasz Index (CHI), are not used for performance evaluation, as ground-truth labels are available for all benchmark datasets and these internal metrics are not suitable for subspace-based algorithms, such as SSC~\cite{elhamifar2013sparse}, DSCN~\cite{dscn}. Instead, these internal metrics are utilized as meta-features in the model selection analysis.
Following the definition in Section~\ref{hard-clustering}, for DBSCAN, a noise-detecting clustering method, the samples identified as noise are treated as a separate cluster when computing the evaluation metrics.
In addition, we provide an empirical comparison of computational cost in Appendix~\ref{appe-time-cost}.
\section{Detailed Information on Algorithms and Datasets}

\subsection{Clustering Algorithms}
\label{appe-algorithms}

\subsubsection{Analysis of Time and Space Complexity}
\label{appe-complexity}

In this section, we detail the time and space complexity of the algorithms evaluated in Table~\ref{tab-time-space-complexity}. Note that the reported time complexity for iteratively optimized algorithms refers to that of a single iteration. The notations used in Table~\ref{tab-time-space-complexity} are defined as follows.

\begin{itemize}
    \item $n$: data size.
    \item $m$: feature dimension.
    \item $k$: number of clusters.
    \item $\rho$: proportion of nonzero entries.
    \item $\Tilde{m}$: $\max\{m,h\}$ where $h$ is the maximal latent dimension of neural networks.
    \item $\theta$: number of neural network parameters.
    \item $p$: dimension of the output from the encoder.
    \item $n_b$: batch size.
    \item $d$: dimension of subspace.
    \item $\Tilde{p}$: $\max\{m,p\}$.
    \item $\Tilde{T}$: maximum number of iterations for the scaling algorithm of P$^2$OT.
    \item $Q$: number of hyperparameter combinations. 
    \item $r$: projected dimension of k-PC.
    
\end{itemize}

\begin{table}[!ht] \tiny
    \caption{Time and space complexity.}
    \label{tab-time-space-complexity}
    \centering
    \resizebox{\columnwidth}{!}{
    \begin{tabular}{l|c|c}
    \toprule
    \textbf{Methods} & \textbf{Time Complexity} & \textbf{Space Complexity} \\
    \midrule
    KMeans~\citep{mcqueen1967some} & $\mathcal{O}(kmn)$ & $\mathcal{O}(nm + km)$ \\
    KernelKMeans~\cite{dhillon2004kernel} &$\mathcal{O}(n^2)$ &$\mathcal{O}(n^2)$ \\
    AggClu~\citep{johnson1967hierarchical} & $\mathcal{O}(n^3)$ & $\mathcal{O}(n^2)$ \\
    DBSCAN~\citep{ester1996density} & $\mathcal{O}(n\log n)$ & $\mathcal{O}(n^2)$ \\
    BIRCH~\citep{zhang1996birch} & $\mathcal{O}(mn)$ & $\mathcal{O}(mn)$ \\
    GMM~\cite{rasmussen1999infinite} & $\mathcal{O}(nkm^2)$ & $\mathcal{O}(nm + km^2)$ \\
    SpeClu~\citep{von2007tutorial} & $\mathcal{O}(mn^2)$ & $\mathcal{O}(n^2)$ \\
    MeanShift~\citep{comaniciu2002mean} & $\mathcal{O}(n^2)$ & $\mathcal{O}(nm)$\\
    SSC~\citep{elhamifar2013sparse} & $\mathcal{O}(mn^2)$ & $\mathcal{O}(mn + \rho n^2)$  \\
    S$^3$COMP-C~\citep{chen2020stochastic} & $\mathcal{O}(m \rho n^3)$ & $\mathcal{O}(mn + \rho n^2)$  \\
    k-PC~\cite{agarwal2004k} & $\mathcal{O}(krn + km\log m)$ & $\mathcal{O}(nm + nr + kr)$\\
    LRR~\cite{liu2012robust} &$\mathcal{O}(mn^2+m^3)$ &$\mathcal{O}(mn+n^2)$\\
    k-FSC~\citep{fan2021large}) & $\mathcal{O}(kdmn)$ & $\mathcal{O}(mn + kmd + kdn)$   \\
    AutoSC~\citep{fan2021large} & $\mathcal{O}(Q(k+m)n^2)$ & $\mathcal{O}(mn + n^2)$\\
    DEC~\citep{dec} & $\mathcal{O}(\Tilde{m}^2n+knp)$ &$\mathcal{O}(\theta+knp+\Tilde{m}n)$\\
    IDEC~\citep{idec} & $\mathcal{O}(\Tilde{m}^2n+knp)$ &$\mathcal{O}(\theta+knp+\Tilde{m}n)$\\
    DSCN~\citep{dscn} & $\mathcal{O}(\Tilde{m}^2n+n^2p)$ &$\mathcal{O}(\theta+n^2+\Tilde{m}n)$ \\
    PICA~\citep{pica} &$\mathcal{O}(k^2n_b+\Tilde{m}^2n_b)$&$\mathcal{O}(\theta+\Tilde{m}n+k^2)$ \\
    ConClu~\citep{cc} &$\mathcal{O}(\Tilde{m}^2n+n^2\Tilde{m})$&$\mathcal{O}(\theta+\Tilde{m}n_b+n_b^2)$ \\
    EDESC~\citep{edesc} & $\mathcal{O}(kdpn + \Tilde{m}\Tilde{p}n)$ & $\mathcal{O}(\theta+\Tilde{m}n + kn + kpd )$\\
    DMICC~\citep{DMICC} &$\mathcal{O}(\Tilde{m}^2n+n^2\Tilde{m})$&$\mathcal{O}(\theta+p^2+n_b^2+n_b\Tilde{m})$ \\
    DIVC~\citep{DIVC} &$\mathcal{O}(k^2n_b+\Tilde{m}^2n_b)$&$\mathcal{O}(\theta+\Tilde{m}n+k^2)$ \\
    P$^2$OT~\cite{zhang2024p} &$\mathcal{O}(\Tilde{m}^2n+\Tilde{T}nm)$&$\mathcal{O}(\theta+\Tilde{m}n)$ \\
    LFSS~\citep{LFSS} &$\mathcal{O}(\Tilde{m}^2n+n^2p)$&$\mathcal{O}(\theta+n_b^2+n_b\Tilde{m})$ \\
    \bottomrule
    \end{tabular}}
\end{table}

\subsubsection{The Search Range of Hyperparameter Configurations (HPCs)}
\label{appe-hpc}
\begin{table*}[!ht]
    \centering
    \caption{The search range of hyperparameter configurations.}
    \label{tab-hpc-search}
    \resizebox{0.9\textwidth}{!}{
    \begin{tabular}{|l|l|c|}
    \hline
    \textbf{Algorithm} & \textbf{HPC (Hyperparameter Configuration)}  & \textbf{\# Total} \\
    \hline
    KMeans &  $ \text{init} \in \{ \text{kmeans++, random}\}$; metric=\{\text{euclidean, manhattan, cosine}\}; n\_init=10; max\_iter=500 & 6 \\
    \hline
    KernelKMeans &  \makecell[l]{\text{kernel}=\text{rbf}; $\text{gamma} \in \{\ 0.01, 0.1, 1.0, 10.0, 100.0 \}$; \\ $\text{init} \in \{ \text{kmeans++, random}\}$; metric=euclidean; max\_iter=500} & 10 \\
    \hline
    AggClu & $\text{metric} \in \{\text{euclidean, manhattan, cosine} \}$; $\text{linkage} \in \{ \text{average, complete, single}\}$ & 9 \\
    \hline
    \multirow{2}{*}{DBSCAN} & $\text{eps} \in \{0.001, 0.005, 0.01, 0.1, 0.2, 0.4, 0.6, 0.8, 1.0, 10.0 \}$; & \multirow{2}{*}{90}\\
     & $\text{min\_sample} \in \{3, 5, 10\}$;  $\text{metric} \in \{\text{euclidean, manhattan, cosine} \}$& \\
     \hline
    BIRCH & $\text{threshold} \in \{0.3, 0.5, 0.7, 0.9\}$; $\text{branching\_factor} \in \{30, 50, 70\}$ & 12\\
    \hline
    GMM &  $\text{covariance\_type} \in \{\text{full, spherical}\}$; $\text{init\_params} \in \{\text{kmeans, kmeans++, random}\}$& 6 \\
    \hline
    \multirow{2}{*}{SpeClu} & affinity=knn; $k = \{3, 5, 10, 20, 30, 50\}$ & \multirow{2}{*}{11}\\
    & affinity=rbf; $\text{gamma} \in \{ 0.1, 0.5, 1.0, 5.0, 10.0\}$& \\
    \hline
    MeanShift & $\text{bandwidth} \in \{ 0.1, 0.3, 0.5, 0.7\}$; $\text{min\_bin\_freq} \in \{1, 3, 5\}$ & 12\\
    \hline
    k-PC & init\_type=k-means; $\text{d} \in \{ 5, 10, 20, 30, 50\}$& 5\\
    \hline
    SSC& $\text{lambda}= \{100.0, 10.0, 1.0, 0.1, 0.01\}$ & 5\\
    \hline
    LRR & $d \in \{5, 10, 20, 50 \}; \lambda \in \{ 0.01, 0.1, 1.0, 10.0 \}$& 16\\
    \hline
    S$^3$COMP-C & $\text{delta} \in \{ 0.1, 0.3\}; \text{lambda} \in \{0.1, 0.3, 0.5\}$ & 6\\
    \hline
    k-FSC&  $d \in \{5, 10, 20\}; \text{lambda}=\{0.01, 0.1, 0.3, 0.5\}$ & 12\\
    \hline
    AutoSC & Auto hyperparameter Search & NA \\
    \hline
    DEC & lr$\in \{1e^{-3}, 1e^{-4}\}$; hidden\_dims $\in\{64,32,16\}$&6\\
    \hline
    IDEC& lr$= 1e^{-4}$; hidden\_dims $\in\{64,32,16\}$; gamma $\in\{0.01,0.1\}$ & 6\\
    \hline
    DSCN & lr$= 1e^{-3}$; hidden\_dims $\in\{64,32,16\}$; $\text{dim\_subspace} \in \{10, 20\}$& 6 \\
    \hline
    PICA & lr$\in \{1e^{-3}, 1e^{-4}\}$; lamda $\in \{0.1,0.5,1\}$&6\\
    \hline
    ConClu & lr$= 1e^{-4}$; instance\_tempature $\in \{0.1,0.5,1.0\}$; cluster\_tempature $\in \{0.1,0.5,1\}$ & 9\\
    \hline
    EDESC &lr$= 1e^{-3}$; beta $\in \{0.1,1,10\}$; d $\in \{1,5,10\}$&9\\
    \hline
    \multirow{2}{*}{DMICC}& lr$= 1e^{-3}$; lamda1 $\in \{1e^{-3},1e^{-4}\};$ lamda2 $\in \{1e^{-3},1e^{-4}\}$&\multirow{2}{*}{12}\\&hidden\_dims $\in\{64,32,16\}$ &\\
    \hline
    DIVC & lr$\in \{1e^{-3}, 1e^{-4}\}$; lamda $\in \{0.1,0.5,1\}$& 6\\
    \hline
    P$^2$OT & rho\_strategy $\in \{\text{sigmoid}, \text{linear} \}$; rho $\in \{0.05, 0.1, 0.2 \}$ & 6 \\
    \hline
    LFSS & lr$= 1e^{-3}$; hidden\_dims $\in \{64, 32, 16\}$; lamda\_da $\in \{0.1, 0.5\}$; temp $= 0.5$ & 6\\
    \hline

    \end{tabular}
    }
\end{table*}

To obtain reliable performance estimates for all algorithms evaluated in CLUBench, we tune the hyperparameter configurations (HPCs) and search for the best performance for each clustering algorithm on each dataset within the predefined search space.
Because the implementation of DBSCAN in \texttt{scikit-learn} does not require $K$ as an input hyperparameter, we tuned its hyperparameter configurations over a broad range, resulting in 90 distinct configurations. This extensive tuning is intended to provide a relatively fair comparison with methods that rely on a predefined number of clusters $K$. The search ranges of hyperparameter configurations for each algorithm are determined based on official recommendations, ablation studies reported in the original papers, or the documentation of widely used third-party libraries (e.g., \texttt{scikit-learn}). The detailed search ranges for all HPCs are summarized in Table~\ref{tab-hpc-search}. It is worth noting that, in Table~\ref{tab-hpc-search}, each value of `eps' in DBSCAN represents a scaling factor. In practice, we define an `eps\_base' as the average pairwise distance among samples and scale it by the specified multipliers to ensure that the `eps' values fall within a reasonable and valid range. For a dataset $\mathcal{X}=\{\mathbf{x}_1, \mathbf{x}_2, \cdots, \mathbf{x}_n\}$, we have 
\begin{equation*}
    \begin{aligned}
        &\text{eps} \leftarrow \text{eps} \times \text{eps\_base},\\
        &\text{eps\_base} = \frac{1}{n(n-1)} \sum_{i=1}^n \sum_{j\neq i}^n \text{dist}(\mathbf{x}_i, \mathbf{x}_j),
    \end{aligned}
\end{equation*}
where $\text{dist}(\cdot, \cdot)$ denotes a distance measure depending on the argument `metric'. Similarly, for `gamma' in SpeClu (Spectral Clustering) and KerKMeans, we have 
\begin{equation*}
    \begin{aligned}
        &\text{gamma} \leftarrow \text{gamma} \times \text{gamma\_base},\\
        &\text{gamma\_base} = \frac{1}{2 \times \text{median}(\{\Vert \mathbf{x}_i - \mathbf{x}_j \Vert_2^2~|~ \mathbf{x}_i, \mathbf{x}_j \in \mathcal{X}, i \neq j \})}.
    \end{aligned}
\end{equation*}

\subsection{Benchmark Datasets}
\label{appe-datasets}

To control the computational scale of the experiments, datasets with a sample size $n>10{,}000$ are uniformly subsampled to $10{,}000$ instances. We report detailed statistics for all datasets used in Table~\ref{tab-all-datasets-1}, Table~\ref{tab-all-datasets-2}, and Table~\ref{tab-all-datasets-3}. 

To characterize the cluster size imbalance ratios of datasets, we report two ratios: 
$r_{mm}:=\frac{\#~\text{samples in minimal cluster}}{\#~\text{samples in maximal cluster}} \in (0, 1]$ and $r_{ma}:=\frac{\#~\text{samples in minimal cluster}}{\#~\text{all samples}} \in (0, 0.5]$. 
These ratios capture the degree of disparity in cluster sizes from complementary perspectives.
In addition, we report an imbalance factor (IR) to quantify the overall degree of class imbalance within a dataset. Specifically, for a dataset containing $K$ classes, we define the class proportion of the $i$-th class as $p_i = \frac{\#~\text{samples in the $i$-th class}}{\sum_{j=1}^{K} \#~\text{samples in the $j$-th class}}$. The imbalance factor is then defined as the standard deviation of the class proportion distribution $\{p_i\}_{i=1}^{K}$, with larger values indicating more severe imbalance.

\begin{table*}[!ht] \tiny
\centering
\caption{Statistics of datasets (1-55).}
\label{tab-all-datasets-1}
\resizebox{\textwidth}{!}{
\begin{tabular}{l|ccccccc}
\toprule
\textbf{Datasets} & \textbf{Type} & \# \textbf{Samples} & \textbf{Dimension} & \# \textbf{clusters} & $\mathbf{r_{mm}}$ & $\mathbf{r_{ma}}$ & \textbf{IR}\\
\midrule
$[1]$ echocardiogram & tabular & 61 & 10 & 2 & 0.386 & 0.279 & 0.221\\
$[2]$ skillcraft1\_master\_table\_dataset& tabular &3303 & 18 & 6 & 0.206 & 0.051 & 0.071\\
$[3]$ breast\_cancer\_wisconsin\_original& tabular &683 & 9 & 2 & 0.538 & 0.350 & 0.150\\
$[4]$ smoker\_condition& tabular &1012 & 7 & 2 & 0.656 & 0.396 & 0.104\\
$[5]$ glass\_identification& tabular &214 & 9 & 6 & 0.118 & 0.042 & 0.127\\
$[6]$ statlog\_image\_segmentation& tabular &2310 & 19 & 7 & 1.000 & 0.143 & 0.000\\
$[7]$ planning\_relax& tabular &182 & 12 & 2 & 0.400 & 0.286 & 0.214\\
$[8]$ customer\_classification& tabular &1000 & 11 & 4 & 0.772 & 0.217 & 0.025\\
$[9]$ pima\_indians\_diabetes\_database& tabular &768 & 8 & 2 & 0.536 & 0.349 & 0.151\\
$[10]$ mobile\_price\_classification& tabular &2000 & 20 & 4 & 1.000 & 0.250 & 0.000\\
$[11]$ spambase& tabular &4601 & 57 & 2 & 0.650 & 0.394 & 0.106\\
$[12]$ rice\_seed\_gonen\_jasmine& tabular &9999 & 10 & 2 & 0.821 & 0.451 & 0.049\\
$[13]$ heart\_attack\_analysis\_prediction\_dataset& tabular &303 & 13 & 2 & 0.836 & 0.455 & 0.045\\
$[14]$ user\_knowledge\_modeling& tabular &258 & 5 & 4 & 0.273 & 0.093 & 0.098\\
$[15]$ world12d& tabular &150 & 12 & 5 & 0.190 & 0.053 & 0.088\\
$[16]$ pumpkin\_seeds& tabular &2500 & 12 & 2 & 0.923 & 0.480 & 0.020\\
$[17]$ iris& tabular &150 & 4 & 3 & 1.000 & 0.333 & 0.000\\
$[18]$ wine& tabular &178 & 13 & 3 & 0.676 & 0.270 & 0.053\\
$[19]$ letter\_recognition& tabular &9992 & 16 & 26 & 0.904 & 0.037 & 0.001\\
$[20]$ mammographic\_mass& tabular &830 & 5 & 2 & 0.944 & 0.486 & 0.014\\
$[21]$ breast\_tissue& tabular &106 & 9 & 6 & 0.636 & 0.132 & 0.028\\
$[22]$ hepatitis& tabular &80 & 19 & 2 & 0.194 & 0.163 & 0.338\\
$[23]$ predicting\_pulsar\_star& tabular &9273 & 8 & 2 & 0.101 & 0.092 & 0.408\\
$[24]$ breast\_cancer\_wisconsin\_prognostic& tabular &569 & 30 & 2 & 0.594 & 0.373 & 0.127\\
$[25]$ wireless\_indoor\_localization& tabular &2000 & 7 & 4 & 1.000 & 0.250 & 0.000\\
$[26]$ date\_fruit& tabular &898 & 34 & 7 & 0.319 & 0.072 & 0.062\\
$[27]$ zoo& tabular &101 & 16 & 7 & 0.098 & 0.040 & 0.118\\
$[28]$ htru2& tabular &9999 & 8 & 2 & 0.101 & 0.092 & 0.408\\
$[29]$ ionosphere& tabular &351 & 34 & 2 & 0.560 & 0.359 & 0.141\\
$[30]$ music\_genre\_classification& tabular &1000 & 26 & 10 & 1.000 & 0.100 & 0.000\\
$[31]$ spectf\_heart& tabular &80 & 44 & 2 & 1.000 & 0.500 & 0.000\\
$[32]$ rice\_dataset\_cammeo\_and\_osmancik& tabular &3810 & 7 & 2 & 0.748 & 0.428 & 0.072\\
$[33]$ ph\_recognition& tabular &653 & 3 & 15 & 0.864 & 0.058 & 0.002\\
$[34]$ banknote\_authentication& tabular &1372 & 4 & 2 & 0.801 & 0.445 & 0.055\\
$[35]$ wine\_quality& tabular &4873 & 11 & 5 & 0.074 & 0.033 & 0.160\\
$[36]$ cardiovascular\_study& tabular &2927 & 15 & 2 & 0.179 & 0.152 & 0.348\\
$[37]$ statlog\_german\_credit& tabular &1000 & 24 & 2 & 0.429 & 0.300 & 0.200\\
$[38]$ boston& tabular &154 & 13 & 3 & 0.371 & 0.169 & 0.121\\
$[39]$ seismic\_bumps& tabular &646 & 24 & 2 & 0.071 & 0.067 & 0.433\\
$[40]$ dry\_bean& tabular &9997 & 16 & 7 & 0.147 & 0.038 & 0.065\\
$[41]$ credit\_risk\_classification& tabular &976 & 11 & 2 & 0.239 & 0.193 & 0.307\\
$[42]$ epileptic\_seizure\_recognition& tabular &5750 & 178 & 5 & 1.000 & 0.200 & 0.000\\
$[43]$ website\_phishing& tabular &1353 & 9 & 3 & 0.147 & 0.076 & 0.188\\
$[44]$ optical\_recognition\_of\_handwritten\_digits& tabular &3823 & 64 & 10 & 0.967 & 0.098 & 0.001\\
$[45]$ siberian\_weather\_stats& tabular &1407 & 11 & 7 & 0.073 & 0.024 & 0.122\\
$[46]$ orbit\_classification\_for\_prediction\_nasa& tabular &1722 & 11 & 3 & 0.065 & 0.056 & 0.371\\
$[47]$ magic\_gamma\_telescope& tabular &9999 & 10 & 2 & 0.542 & 0.352 & 0.148\\
$[48]$ raisin& tabular &900 & 7 & 2 & 1.000 & 0.500 & 0.000\\
$[49]$ patient\_treatment\_classification& tabular &4412 & 10 & 2 & 0.679 & 0.404 & 0.096\\
$[50]$ fetal\_health\_classification& tabular &2126 & 21 & 3 & 0.106 & 0.083 & 0.316\\
$[51]$ dermatology& tabular &358 & 34 & 6 & 0.180 & 0.056 & 0.373\\
$[52]$ secom& tabular &1567 & 590 & 2 & 0.071 & 0.066 & 0.000\\
$[53]$ paris\_housing\_classification& tabular &10000 & 17 & 2 & 0.145 & 0.127 & 0.053\\
$[54]$ seeds& tabular &210 & 7 & 3 & 1.000 & 0.333 & 0.275\\
$[55]$ wine\_customer& tabular &178 & 13 & 3 & 0.676 & 0.270 & 0.000\\

\bottomrule
\end{tabular}}
\label{tab-dataset-detail-1}
\end{table*}

\begin{table*}[!ht] \tiny
\centering
\caption{Statistics of datasets (56-110).}
\label{tab-all-datasets-2}
\resizebox{\textwidth}{!}{
\begin{tabular}{l|ccccccc}
\toprule
\textbf{Datasets} & \textbf{Type} & \# \textbf{Samples} & \textbf{Dimension} & \# \textbf{clusters} & $\mathbf{r_{mm}}$ & $\mathbf{r_{ma}}$ & \textbf{IR}\\
\midrule

$[56]$ crowdsourced\_mapping& tabular &9997 & 28 & 4 & 0.060 & 0.043 & 0.212\\
$[57]$ durum\_wheat\_features& tabular &9000 & 236 & 3 & 1.000 & 0.333 & 0.093\\
$[58]$ classification\_in\_asteroseismology& tabular &1001 & 3 & 2 & 0.404 & 0.288 & 0.063\\
$[59]$ birds\_bones\_and\_living\_habits& tabular &413 & 10 & 6 & 0.185 & 0.056 & 0.000\\
$[60]$ microbes& tabular &9995 & 24 & 10 & 0.082 & 0.020 & 0.097\\
$[61]$ image\_segmentation& tabular &210 & 19 & 7 & 1.000 & 0.143 & 0.440\\
$[62]$ water\_quality& tabular &2011 & 9 & 2 & 0.676 & 0.403 & 0.235\\
$[63]$ insurance\_company\_benchmark& tabular &5822 & 85 & 2 & 0.064 & 0.060 & 0.115\\
$[64]$ harbermans\_survival& tabular &306 & 3 & 2 & 0.360 & 0.265 & 0.175\\
$[65]$ yeast& tabular &1459 & 8 & 8 & 0.065 & 0.021 & 0.132\\
$[66]$ heart\_disease& tabular &297 & 13 & 5 & 0.081 & 0.044 & 0.004\\
$[67]$ ecoli& tabular &327 & 7 & 5 & 0.140 & 0.061 & 0.052\\
$[68]$ extyaleb& tabular &319 & 30 & 5 & 0.954 & 0.194 & 0.171\\
$[69]$ breast\_cancer\_coimbra& tabular &116 & 9 & 2 & 0.812 & 0.448 & 0.061\\
$[70]$ student\_grade& tabular &395 & 29 & 2 & 0.491 & 0.329 & 0.234\\
$[71]$ human\_stress\_detection& tabular &2001 & 3 & 3 & 0.634 & 0.250 & 0.004\\
$[72]$ fraud\_detection\_bank& tabular &9999 & 112 & 2 & 0.362 & 0.266 & 0.031\\
$[73]$ pen\_based\_recognition\_of\_handwritten\_digits& tabular &7494 & 16 & 10 & 0.922 & 0.096 & 0.000\\
$[74]$ diabetic\_retinopathy\_debrecen& tabular &1151 & 19 & 2 & 0.884 & 0.469 & 0.026\\
$[75]$ pistachio& tabular &2148 & 28 & 2 & 0.744 & 0.426 & 0.262\\
$[76]$ turkish\_music\_emotion& tabular &400 & 50 & 4 & 1.000 & 0.250 & 0.000\\
$[77]$ parkinsons& tabular &195 & 22 & 2 & 0.327 & 0.246 & 0.000\\
$[78]$ weather& tabular &365 & 192 & 7 & 0.603 & 0.121 & 0.148\\
$[79]$ blood\_transfusion\_service\_center& tabular &748 & 4 & 2 & 0.312 & 0.238 & 0.004\\
$[80]$ mfeat-karhunen& tabular &2000 & 64 & 10 & 1.000 & 0.100 & 0.039\\
$[81]$ mfeat-factors& tabular &2000 & 216 & 10 & 1.000 & 0.100 & 0.116\\
$[82]$ wall-robot-navigation& tabular &5456 & 24 & 4 & 0.149 & 0.060 & 0.007\\
$[83]$ Waveform& tabular &5000 & 21 & 3 & 0.971 & 0.329 & 0.053\\
$[84]$ gas-drift& tabular &10000 & 128 & 6 & 0.546 & 0.118 & 0.005\\
$[85]$ mfeat-morphological& tabular &2000 & 6 & 10 & 1.000 & 0.100 & 0.000\\
$[86]$ JapaneseVowels& tabular &9961 & 14 & 9 & 0.485 & 0.079 & 0.124\\
$[87]$ rmftsa\_sleepdata& tabular &1024 & 2 & 4 & 0.233 & 0.092 & 0.337\\
$[88]$ first-order-theorem-proving& tabular &6118 & 51 & 6 & 0.190 & 0.079 & 0.062\\
$[89]$ gina\_prior2& tabular &3468 & 784 & 10 & 0.822 & 0.091 & 0.153\\
$[90]$ fabert& tabular &8237 & 800 & 7 & 0.261 & 0.061 & 0.064\\
$[91]$ dilbert& tabular &10000 & 2000 & 5 & 0.934 & 0.191 & 0.000\\
$[92]$ synthetic\_control& tabular &600 & 60 & 6 & 1.000 & 0.167 & 0.009\\
$[93]$ Drug Consumption& tabular &1749 & 12 & 4 & 0.261 & 0.113 & 0.053\\
$[94]$ shuttle& tabular &10000 & 9 & 2 & 0.195 & 0.163 & 0.005\\
$[95]$ tr45.wc& tabular &676 & 8261 & 9 & 0.113 & 0.027 & 0.000\\
$[96]$ steel-plates-fault& tabular &1941 & 33 & 2 & 0.531 & 0.347 & 0.000\\
$[97]$ fbis.wc& tabular &2196 & 2000 & 11 & 0.128 & 0.030 & 0.000\\
$[98]$ mfeat-fourier& tabular &2000 & 76 & 10 & 1.000 & 0.100 & 0.000\\
$[99]$ vehicle& tabular &846 & 18 & 4 & 0.913 & 0.235 & 0.000\\
$[100]$ micro-mass& tabular & 360 & 1300 & 10 & 1.000 & 0.100 & 0.000\\
$[101]$ ISOLET& tabular &7797 & 617 & 26 & 0.993 & 0.038 & 0.000\\
$[102]$ poker-hand& tabular &10000 & 10 & 2 & 0.843 & 0.457 & 0.000\\
$[103]$ tamilnadu-electricity& tabular &10000 & 2 & 20 & 0.480 & 0.030 & 0.000\\
$[104]$ mnist64& image &1082 & 64 & 6 & 0.967 & 0.164 & 0.000\\
$[105]$ MNIST\_CLIP$^+$& image &9996 & 512 & 10 & 0.801 & 0.090 & 0.000\\
$[106]$ fashion\_mnist& image &3000 & 784 & 10 & 1.000 & 0.100 & 0.001\\
$[107]$ FashionMNIST\_CLIP$^+$& image &10000 & 512 & 10 & 1.000 & 0.100 & 0.038\\
$[108]$ cifar10& image &3250 & 1024 & 10 & 1.000 & 0.100 & 0.021\\
$[109]$ CIFAR10\_CLIP$^+$& image &10000 & 512 & 10 & 1.000 & 0.100 & 0.153\\
$[110]$ coil20$^*$& image &1440 & 400 & 20 & 1.000 & 0.050 & 0.062\\$[111]$ COIL20\_CLIP$^+$& image &1440 & 512 & 20 & 1.000 & 0.050 & 0.000\\
\bottomrule
\end{tabular}}
\label{tab-dataset-detail-2}
\end{table*}

\begin{table*}[!ht] \tiny
\centering
\caption{Statistics of datasets (111-131).}
\label{tab-all-datasets-3}
\resizebox{\textwidth}{!}{
\begin{tabular}{l|ccccccc}
\toprule
\textbf{Datasets} & \textbf{Type} & \# \textbf{Samples} & \textbf{Dimension} & \# \textbf{clusters} & $\mathbf{r_{mm}}$ & $\mathbf{r_{ma}}$ & \textbf{IR}\\
\midrule
$[112]$ labeled\_faces\_in\_the\_wild& image &2200 & 5828 & 2 & 1.000 & 0.500 & 0.006\\
$[113]$ flickr\_material\_database& image &997 & 1536 & 10 & 0.990 & 0.099 & 0.053\\
$[114]$ street\_view\_house\_numbers& image &732 & 1024 & 10 & 0.341 & 0.064 & 0.000\\
$[115]$ har& image &735 & 561 & 6 & 0.702 & 0.135 & 0.006\\
$[116]$ Indian\_pines& image &8858 & 220 & 5 & 0.121 & 0.055 & 0.000\\
$[117]$ satellite\_image& image &6435 & 36 & 6 & 0.408 & 0.097 & 0.000\\
$[118]$ olivetti\_faces& image &400 & 4096 & 40 & 1.000 & 0.025 & 0.000\\
$[119]$ cnae9& text &1080 & 856 & 9 & 1.000 & 0.111 & 0.000\\
$[120]$ imdb& text &3250 & 700 & 2 & 1.000 & 0.500 & 0.000\\
$[121]$ hate\_speech& text &3221 & 100 & 3 & 0.075 & 0.058 & 0.000\\
$[122]$ sentiment\_labeld\_sentences& text &2748 & 200 & 2 & 0.983 & 0.496 & 0.000\\
$[123]$ sms\_spam\_collection& text &835 & 500 & 2 & 0.155 & 0.134 & 0.000\\
$[124]$ wos& text &9997 & 4096 & 7 & 0.223 & 0.069 & 0.000\\
$[125]$ enron& text &9999 & 4096 & 2 & 0.990 & 0.497 & 0.315\\
$[126]$ reuters& text &6576 & 4096 & 3 & 0.562 & 0.243 & 0.004\\
$[127]$ 20newsgroups& text &9991 & 4096 & 20 & 0.612 & 0.033 & 0.366\\
$[128]$ Mouse\_retina& tabular (BioInfo) &8352 & 6198 & 5 & 0.054 & 0.043 & 0.073\\
$[129]$ Campbell& tabular(BioInfo) &9993 & 26774 & 14 & 0.052 & 0.024 & 0.003\\
$[130]$ PCam& image &4000 & 27648 & 2 & 0.977 & 0.494 & 0.302\\
$[131]$ Baron Human& tabular (BioInfo) &8451 & 20125 & 9 & 0.069 & 0.020 & 0.111\\
\bottomrule
\end{tabular}}
\label{tab-dataset-detail-3}
\end{table*}

\section{More Experimental Results and Analysis}

\subsection{Overall Performance Analysis}
\label{appe-overall-analysis}

\begin{figure}[!ht]
    \centering
    \includegraphics[width=\linewidth]{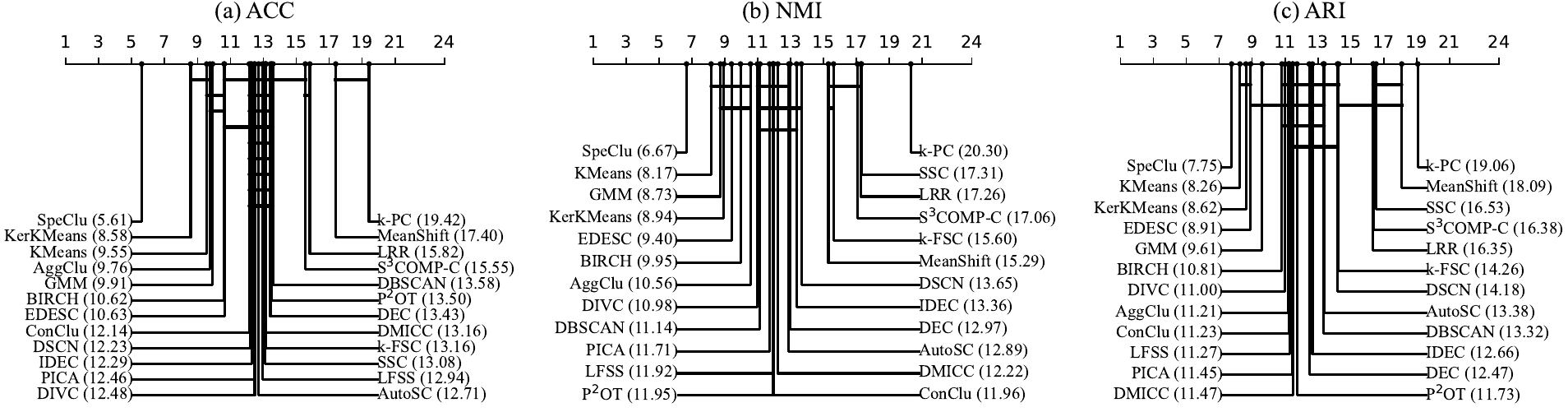}
    \caption{The CD diagram of best performance on all datasets by paired t-test. The `(value)' entries are the average ranks.}
    \label{fig-cd-diagram}
\end{figure}

\paragraph{Statistical Test}
Figure~\ref{fig-cd-diagram} illustrates the statistical performance differences among all algorithms using paired t-tests, where `(value)' denotes the average rank of the corresponding algorithm. SpeClu is statistically indistinguishable from a small subset of top-performing methods, as indicated by the non-overlapping CD intervals. This demonstrates that SpeClu not only attains strong absolute performance but also exhibits stable superiority across diverse datasets and three evaluation metrics.

\begin{figure}[!ht]
    \centering
    \includegraphics[width=\linewidth]{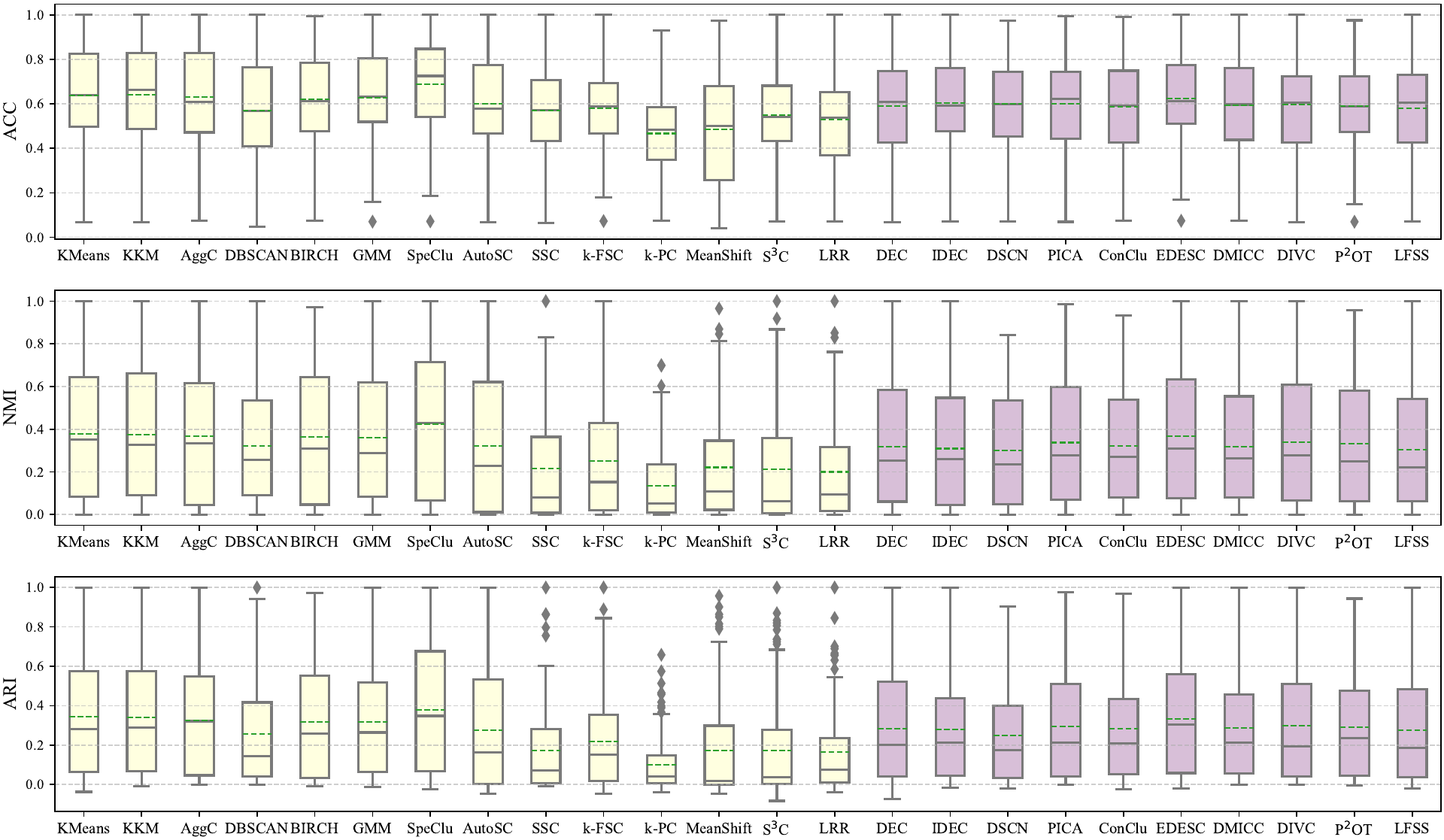}
    \caption{The performance distributions of each algorithm across all datasets under best hyperparameter configurations. The boxplots with \textcolor{yellow!70}{light-yellow} and \textcolor{purple!50}{purple} correspond to the conventional and deep clustering algorithms, respectively. The \textcolor{green!50}{green dashed line} denotes the average performance.}
    \label{fig-box-plot}
\end{figure}

\paragraph{Performance Distributions}
Beyond the average performance reported in Table~\ref{tab-all-avg-p}, Figure~\ref{fig-box-plot} illustrates the performance distributions of each algorithm across the 131 datasets under different evaluation metrics. In the figure, `KKM', `AggC', and `S$^3$C' denote Kernel KMeans, AggClu, and S$^3$COMP-C, respectively. As shown in Figure~\ref{fig-box-plot}, all evaluated methods exhibit similar performance variability, ranging from high (strong performance) to low (poor performance) values across datasets. This observation suggests that, on the one hand, algorithms achieving strong average performance may still be suboptimal on specific datasets. On the other hand, the wide performance distributions reflect the diversity of the collected datasets and indicate that the proposed benchmark provides a meaningful and challenging testbed for evaluating clustering algorithms.

\begin{figure}[!ht]
    \centering
    \includegraphics[width=0.95\linewidth]{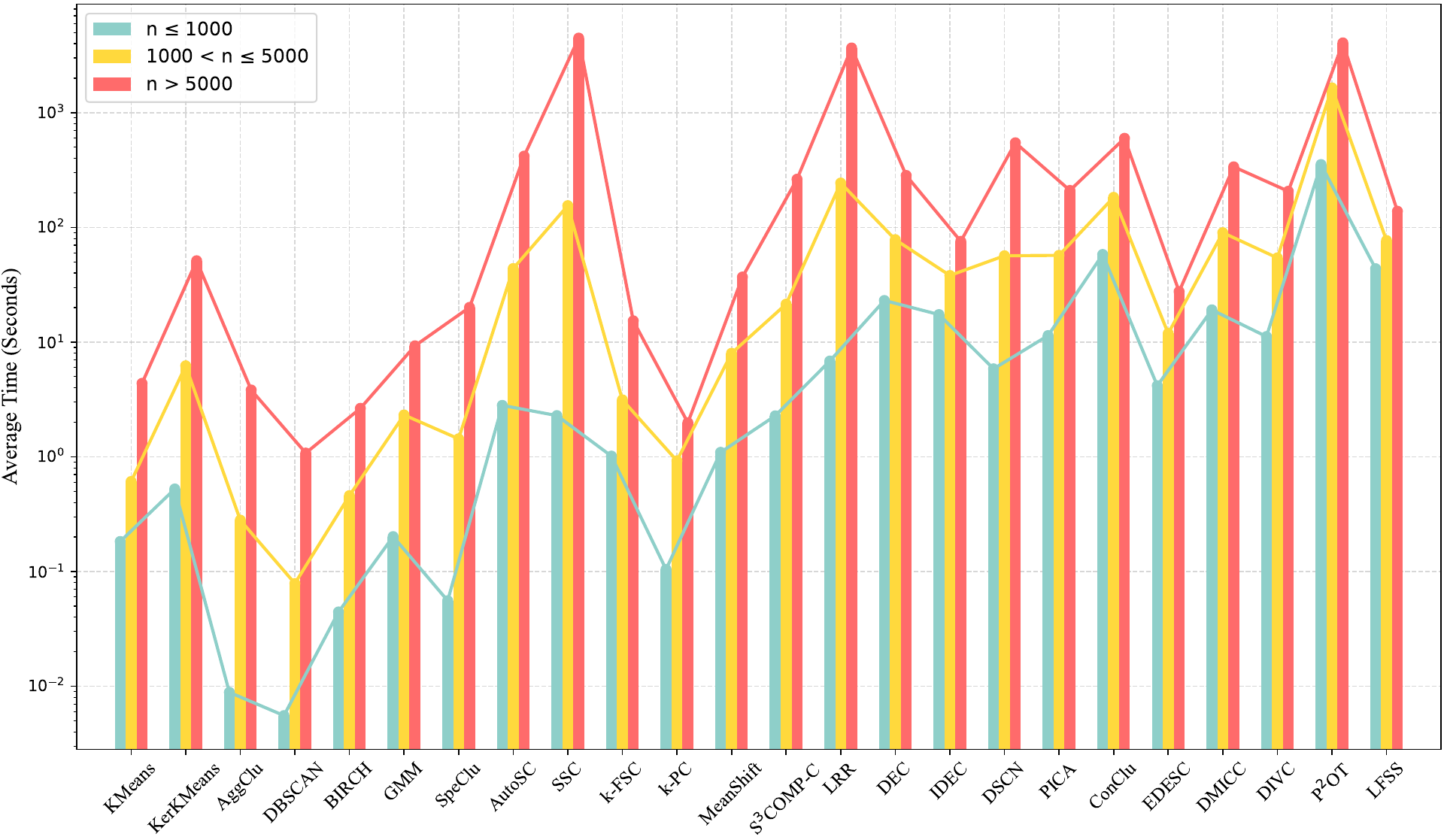}
    \caption{The average implementation time (seconds) across datasets with feature dimension < 500. The $n$ denotes the sample size.}
    \label{fig-avg-time-1}
\end{figure}

\begin{figure}[!ht]
    \centering
    \includegraphics[width=0.95\linewidth]{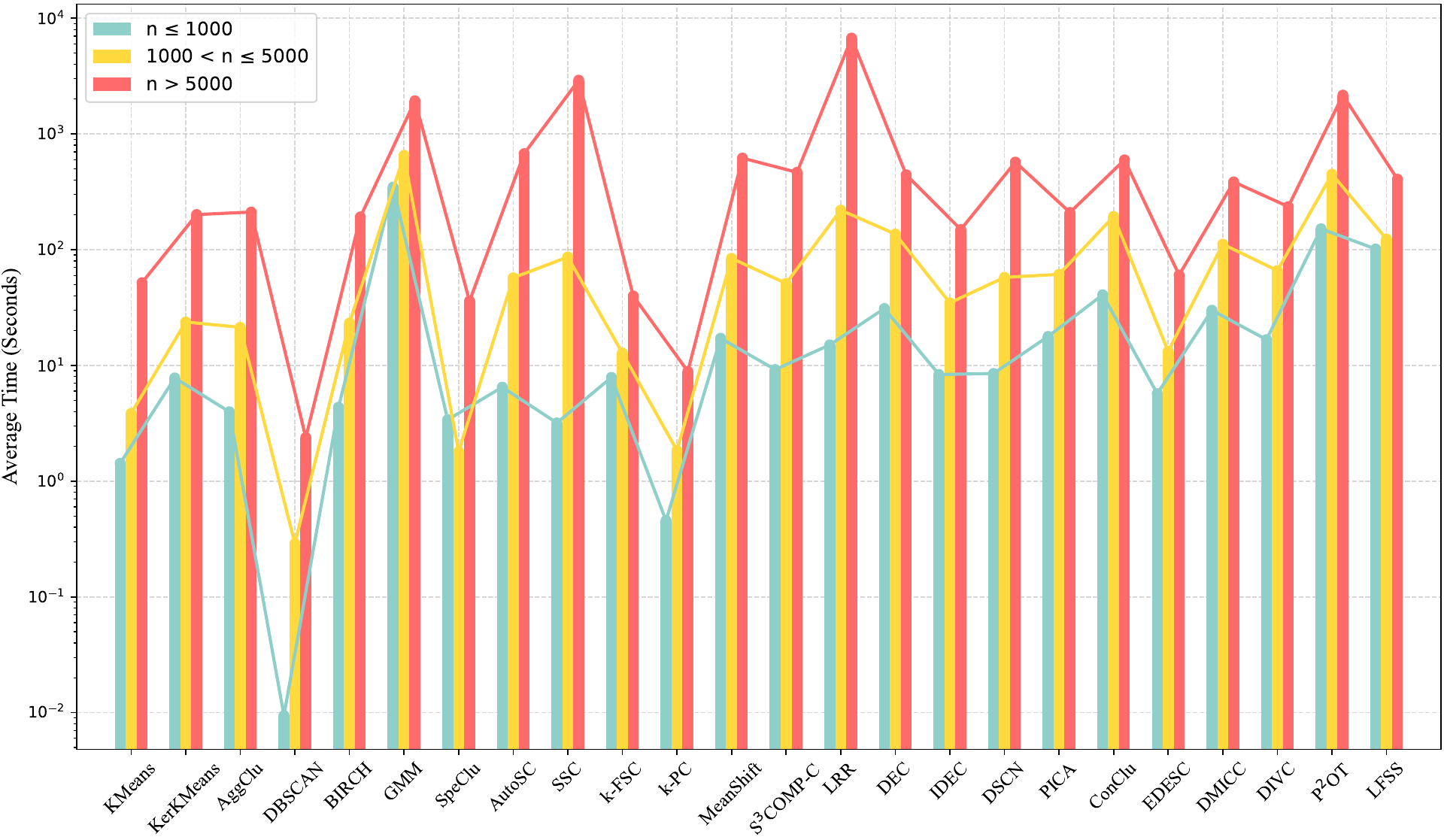}
    \caption{The average implementation time (seconds) across datasets with feature dimension $\geq$ 500. The $n$ denotes the sample size.}
    \label{fig-avg-time-2}
\end{figure}

\paragraph{Empirical Computational Cost}
\label{appe-time-cost}
To intuitively observe the empirical computation cost of each algorithm, we visualize the average running time of all algorithms in Figure~\ref{fig-avg-time-1} and Figure~\ref{fig-avg-time-2}. Specifically, Figure~\ref{fig-avg-time-1} represents results computed on datasets with feature dimension $m < 500$, while Figure~\ref{fig-avg-time-2} reports results on datasets with feature dimension $m \geq 500$. This separation facilitates clearer visualization and comparison of computational differences across algorithms. From the two figures, among conventional clustering algorithms, SSC and LRR exhibit relatively high average computational cost, whereas DBSCAN and k-PC demonstrate substantially lower average running times. For deep clustering methods, P$^2$OT incurs significantly higher computational cost than all other deep clustering baselines, indicating a considerable scalability challenge in practice.

\subsection{FM-based Clustering Analysis}
\label{appe-fm}

Table~\ref{tab-image-datasets} and Table~\ref{tab-text-datasets} summarize the statistical information of image and text datasets used in Section~\ref{analysis-fm}. The prompt template designed for in-context learning of tabular clustering tasks is presented in Table \ref{tab:prompt-template}. Specifically, the three models evaluated on tabular datasets are \texttt{llama-4-maverick}, \texttt{DeepSeek-V3.2}, and \texttt{gpt-5-mini-2025-\\08-07}. For image clustering, we compare several top-performing baselines (identified in Section~\ref{analysis-1}) operating on these feature embeddings with state-of-the-art image-based deep clustering methods~\citep{dec, edesc, LFSS, li2025you}. The overall results are summarized in Table~\ref{tab-image-performance-2}.

\begin{table}[!ht]
    \caption{A summary of image datasets.}
    \label{tab-image-datasets}
    \centering
    \resizebox{0.6\columnwidth}{!}{
    \begin{tabular}{|c|c|c|c|c|c|}
    \hline
    \textbf{Datasets} & STL-10 & CIFAR-10 & CIFAR-20 & ImageNet-10 & ImageNet-Dogs \\
    \hline
    \# \textbf{Samples} & 13,000 & 60,000 & 60,000 & 13,000 & 19,5000 \\
    \# \textbf{Clusters} & 10 & 10 & 20 & 10 & 15\\
    \textbf{Image Size} & 96$\times$96$\times$3 & 32$\times$32$\times$3 & 32$\times$32$\times$3 & 96$\times$96$\times$3 & 96$\times$96$\times$3\\
    \hline
    \end{tabular}
    }
\end{table}

\begin{table}[!ht]
    \caption{A summary of text datasets.}
    \label{tab-text-datasets}
    \centering
    \resizebox{0.6\columnwidth}{!}{
    \begin{tabular}{|c|c|c|c|c|c|}
    \hline
    \textbf{Datasets} & 20Newsgroups & Enron & IMDB & Reuters21578 & WOS \\
    \hline
    \# \textbf{Samples} & 9,991 & 9,999 & 9,999 & 6,576 & 9,997 \\
    \# \textbf{Clusters} & 20 & 2 & 2 & 3 & 7\\
    \hline
    \end{tabular}
    }
\end{table}

\begin{figure}[!ht]
    \centering
   \includegraphics[width=0.8\linewidth]{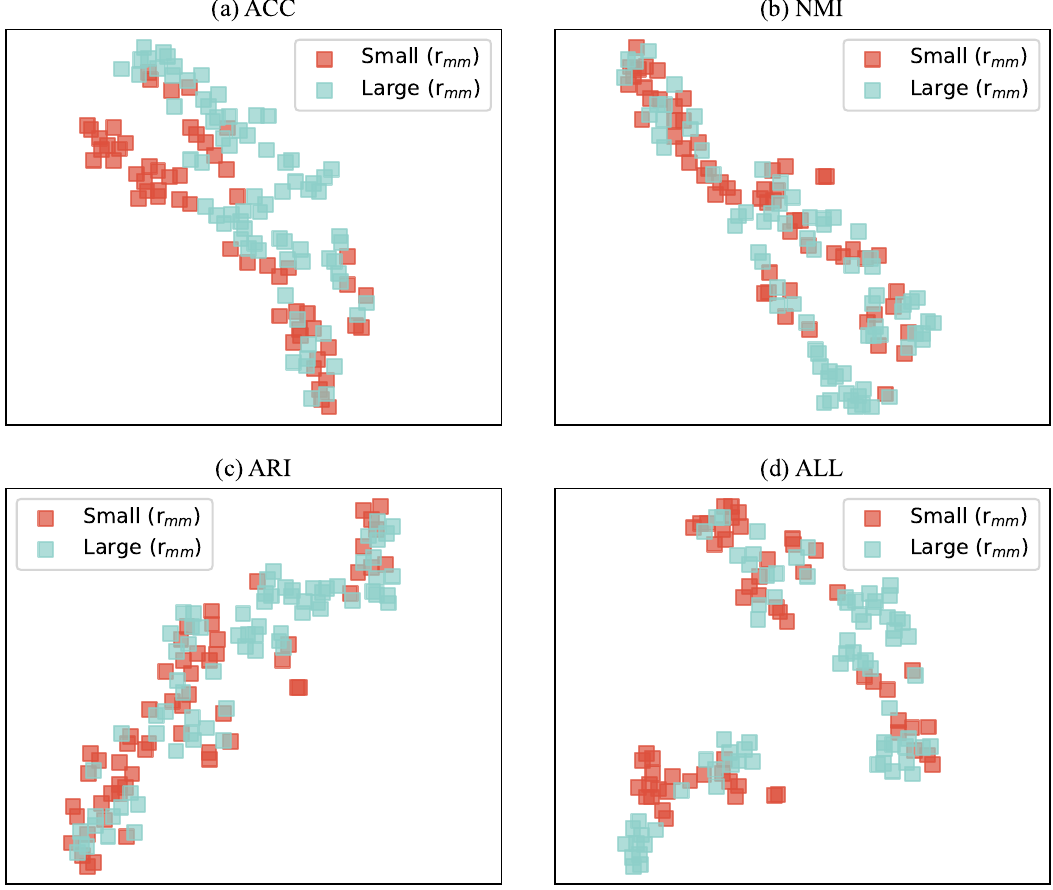}
     \caption{The t-SNE visualization results on dataset performance vectors. }
    \label{fig-tsne-datasets}
\end{figure}

\begin{table*}[!ht]
\centering
\caption{Clustering comparison on image datasets with different pretrained representations. The $*$ indicates the performance values reported in the original paper,  which were obtained using the original image data. The best performances on each dataset are marked in \textcolor{orange}{\textbf{bold}}.} 
\label{tab-image-performance-2}
\resizebox{\textwidth}{!}{
\begin{tabular}{l|ccc|ccc|ccc|ccc|ccc|ccc}
\toprule
& \multicolumn{3}{c|}{\textbf{STL-10}} & \multicolumn{3}{c|}{\textbf{CIFAR-10}} & \multicolumn{3}{c|}{\textbf{CIFAR-20}} & \multicolumn{3}{c|}{\textbf{ImageNet-10 }} &\multicolumn{3}{c|}{\textbf{ImageNet-Dogs}} &\multicolumn{3}{c}{\textbf{Avg.}}\\
\cmidrule(lr){2-4}\cmidrule(lr){5-7}\cmidrule(lr){8-10}\cmidrule(lr){11-13}\cmidrule(lr){14-16} \cmidrule(lr){17-19}
{Method}
& NMI & ACC & ARI & NMI & ACC & ARI
& NMI & ACC & ARI & NMI & ACC & ARI
& NMI & ACC & ARI & NMI & ACC & ARI \\
\midrule
DEC$^*$ (ICML2016)  & 27.6 & 35.9 & 18.6 & 25.7 & 30.1 & 16.1 & 13.6 & 18.5 & 5.0  & 28.2 & 38.1 & 20.3 & 12.2 & 19.5 & 7.9 & 21.4 & 28.4 & 13.5\\
EDESC$^*$ (CVPR2022)& 68.7 & 74.5 & - & 46.4 & 62.7 & - & 37.0 & 38.5 & - & - & - & - & - & - & - & 50.7 & 58.5 & -\\
LFSS$^*$ (ICML2025)& 77.1 & 86.1 & 74.0 & 87.2 & 93.4 & 86.6 & 59.9 & 58.7 & 43.5 & 85.6 & 93.2 & 85.7 & 61.7 & 69.1 & 53.3 & 74.3 & 80.1 & 68.6\\
DCBoost$^*$ (NeurIPS2025)& \textbf{86.7} & \textbf{93.6} & \textbf{86.6} &
\cellcolor{orange!40}\textbf{91.1} & \cellcolor{orange!40}\textbf{96.0} & \cellcolor{orange!40}\textbf{91.6} &
\cellcolor{orange!40}\textbf{64.5} & \cellcolor{orange!40}\textbf{63.9} & \cellcolor{orange!40}\textbf{49.2} &
\textbf{92.7} & \textbf{97.1} & \textbf{93.7} &
\textbf{76.3} & \textbf{79.7} & \textbf{70.7} &
\cellcolor{orange!40}\textbf{82.3} & \cellcolor{orange!40}\textbf{86.1} & \cellcolor{orange!40}\textbf{78.4} \\
\midrule
DEC (ResNet18) & 63.6 & 65.2 & 47.3 & \textbf{58.4} & 64.2 & \textbf{48.4} & 36.0 & 38.3 & 20.5 & 89.1 & 86.8 & 82.9 & 67.3 & 60.4 & 48.7 & 62.8 & 62.9 & 49.5\\
EDESC (ResNet18) & \textbf{84.9} & \textbf{90.2} & \textbf{83.1} & 54.6 & \textbf{66.2} & 45.9 & \textbf{39.9} & 39.7 & \textbf{25.3} & \textbf{96.4} & \textbf{98.6} & \textbf{97.0} & 77.3 & 79.4 & 68.7 & \textbf{70.6} & 74.8 & \textbf{64.0}\\
LFSS (ResNet18) & 77.8 & 88.1 & 75.9 & 50.8 & 65.6 & 44.1 & 36.5 & 40.6 & 22.5 & 94.2 & 97.7 & 95.1 & 82.3 & \textbf{89.9} & \textbf{80.2} & 68.3 & \textbf{76.3} & 63.5\\
KMeans (ResNet18) & 82.7 & 89.9 & 79.2 & 52.1 & 64.0 & 41.0 & 38.0 & \textbf{41.0} & 22.9 & 96.4 & 98.5 & 96.9 & 81.6 & 80.1 & 74.0 & 67.1 & 71.8 & 58.6\\
SpeClu (ResNet18) & 77.9 & 86.7 & 73.0 & 43.7 & 55.5 & 31.7 & 35.1 & 38.4 & 20.5 & 95.9 & 98.4 & 96.6 & \textbf{83.1} & 80.2 & 71.4 & 67.1 & 71.8 & 58.6\\
\midrule
DEC (ResNet50) & 66.3 & 66.6 & 51.7 & 62.9 & 69.0 & 52.4 & 43.1 & 44.2 & 27.3 & 97.4 & 98.9 & 97.7 & 88.0 & 87.6 & 82.1 & 71.5 & 73.2 & 62.2\\
EDESC (ResNet50) & \textbf{91.3} & \textbf{95.9} & \textbf{91.3} & \textbf{64.5} & 69.7 & \textbf{56.7} &
\textbf{45.7} & 43.7 & \textbf{30.1} &
\cellcolor{orange!40}\textbf{99.0} & \cellcolor{orange!40}\textbf{99.6} & \cellcolor{orange!40}\textbf{99.2} &
90.5 & 91.8 & 86.3 &
\textbf{78.2} & 80.1 & \textbf{72.7}\\
LFSS (ResNet50) & 75.5 & 87.5 & 74.6 & 54.8 & 67.5 & 47.3 & 40.4 & 43.6 & 25.9 & 95.4 & 98.3 & 96.3 & 89.1 & 94.6 & 88.9 & 71.0 & 78.3 & 66.6\\
KMeans (ResNet50) & 87.6 & 93.5 & 86.3 & 56.9 & \textbf{70.2} & 48.8 & 42.9 & \textbf{46.3} & 27.8 & 98.4 & 99.4 & 98.7 & 92.1 & 95.9 & 91.4 & 75.5 & \textbf{81.0} & 70.6\\
SpeClu (ResNet50) & 86.9 & 92.1 & 83.3 & 51.7 & 65.0 & 42.0 & 38.6 & 41.3 & 23.9 & 98.5 & 99.4 & 98.8 &
\cellcolor{orange!40}\textbf{92.4} & \cellcolor{orange!40}\textbf{96.0} & \cellcolor{orange!40}\textbf{91.7} &
73.6 & 78.7 & 67.9\\
\midrule
DEC (CLIP) & 79.9 & 71.8 & 62.6 & 75.4 & 79.6 & 68.3 & 55.2 & 51.0 & 34.9 & 94.5 & 97.2 & 94.0 & 39.9 & 41.5 & 26.2 & 68.9 & 68.2 & 57.2\\
EDESC (CLIP) & 95.6 & 98.2 & 96.2 & \textbf{83.0} & 84.6 & 77.9 & \textbf{56.4} & \textbf{53.5} & \textbf{38.7} & \textbf{98.5} & \textbf{99.4} & \textbf{98.8} & 49.4 & 40.9 & 30.9 & \textbf{76.5} & 75.3 & 68.5\\
LFSS (CLIP) & 94.7 & 97.8 & 95.2 & 82.6 & \textbf{91.4} & \textbf{82.0} & 53.4 & 54.1 & 38.1 & 97.0 & 98.9 & 97.5 & 49.7 & 46.1 & 34.6 & 75.4 & \textbf{77.6} & \textbf{69.5}\\
KMeans (CLIP) & 95.1 & 98.0 & 95.7 & 78.7 & 86.6 & 70.6 & 52.7 & 52.2 & 34.3 & 97.5 & 99.0 & 97.9 & \textbf{50.9} & \textbf{51.4} & \textbf{35.7} & 74.9 & 77.4 & 66.8\\
SpeClu (CLIP) &
\cellcolor{orange!40}\textbf{96.3} & \cellcolor{orange!40}\textbf{98.5} & \cellcolor{orange!40}\textbf{96.8} &
79.0 & 85.4 & 67.3 &
47.3 & 46.4 & 28.8 &
98.2 & 99.4 & 98.6 &
48.4 & 51.6 & 35.1 &
73.8 & 76.2 & 65.3\\
\bottomrule
\end{tabular}}
\end{table*}

\begin{table}[!ht]
\centering
\caption{Prompt template used in FM-based tabular clustering}
\label{tab:prompt-template}

\begin{tcolorbox}[
  enhanced,
  title={Prompt template used in FM-based tabular clustering},
  colback=gray!2,
  colframe=black!40,
  boxrule=0.6pt,
  arc=2mm,
  left=2mm,right=2mm,top=1.2mm,bottom=1.2mm,
]
\begin{lstlisting}[style=prompt]
System:
You are a clustering engine.

You MUST output strict json.

Output requirements (MUST follow):
1) Output ONLY a JSON object with keys: "N", "K", "labels".
2) "N" must equal {N}. "K" must equal {K}.
3) "labels" must be a list of length N={N}.
4) Each label must be an integer in {{0,1,...,K-1}} i.e. [0, {K-1}].
5) Every label in {{0,...,K-1}} must appear at least once (use exactly K clusters).
6) No extra keys, no explanation, no markdown, no code.
7) Before you output the final JSON, internally verify:
    a) len(labels)=={N}.
    b) set(labels)=={0,1,...,K-1}.
    c) all labels are ints.
8) If any check fails, fix labels and re-check until all pass.

EXAMPLE JSON OUTPUT:
{{"N": 5, "K": 2, "labels": [0, 1, 1, 0, 1]}}

User:
Return the results as JSON.
N={N}, D={D}, K={K}
X (row0 first):
{Textualized Matrix Rows}

\end{lstlisting}
\end{tcolorbox}

\end{table}

\subsection{Similarity Analysis of Datasets}
\label{appe-simi-data}

\paragraph{Among Datasets}
Following a procedure analogous to that used for algorithm-level analysis, we can obtain an overall performance matrix for all datasets as $\Tilde{\mathbf{P}}_\text{all} = [\mathbf{P}_\text{acc}; \mathbf{P}_\text{nmi}; \mathbf{P}_\text{ari}] \in \mathbb{R}^{N \times 3M}$, where each row $[\Tilde{\mathbf{P}}_\text{all}]_{i,:}$ represents the performance of the $i$-th dataset across all algorithms and evaluation metrics. Applying t-SNE~\cite{maaten2008visualizing} to these dataset-level performance vectors yields a two-dimensional visualization in Figure~\ref{fig-tsne-datasets} (d) that captures similarities among datasets in terms of their clustering behavior. In addition to the aggregated representation, we further apply t-SNE separately to $\mathbf{P}_{\text{acc}}$, $\mathbf{P}_{\text{nmi}}$, and $\mathbf{P}_{\text{ari}}$, with the corresponding visualizations shown in Figure~\ref{fig-tsne-datasets} (a)-(c). In Figure~\ref{fig-tsne-datasets}, datasets are divided into two groups according to the value of $r_{mm}$ (threshold=0.5), which measures the degree of cluster size imbalance. These separation patterns in plots (a)-(c) are significantly distinct, where the metric ACC shows a stronger correlation with $r_{mm}$ than NMI and ARI.

\subsection{Low-rank Analysis}
\label{appe-low-rank}

\begin{figure}[!ht]
    \centering
    \includegraphics[width=\linewidth]{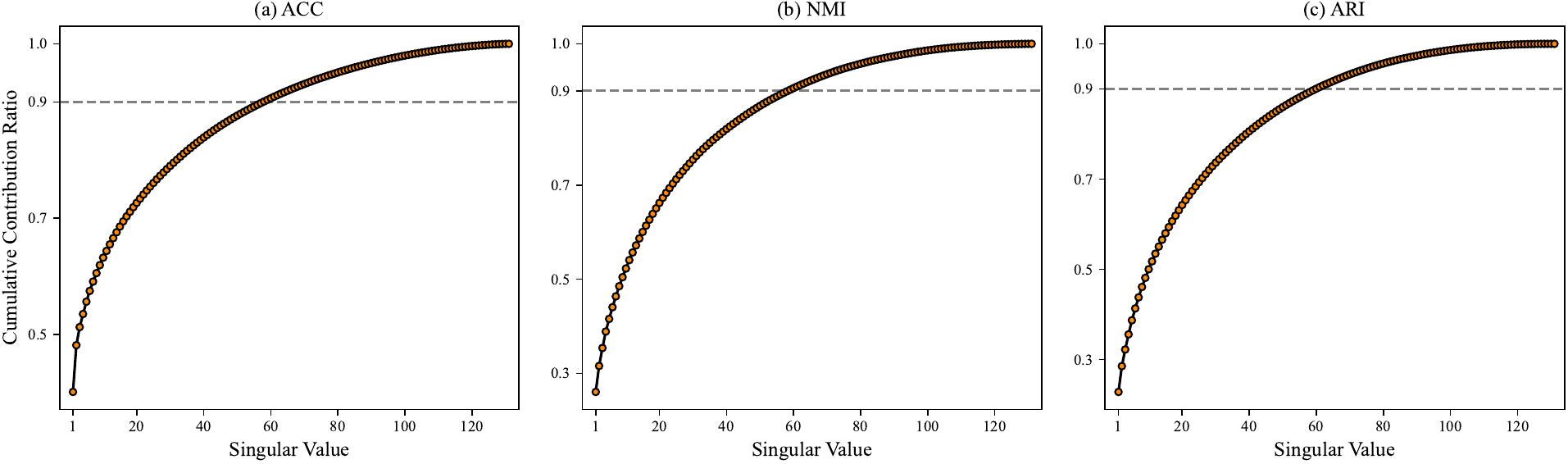}
    \caption{Cumulative contribution ratio of singular values of performance matrices $\mathbf{P}_{acc}, \mathbf{P}_{nmi}, \mathbf{P}_{ari}$.}
    \label{fig-singular-ccr}
\end{figure}
We also analyze the low-rank structure of performance matrices $\mathbf{\hat{P}}_{acc}, \mathbf{\hat{P}}_{nmi}$ and $\mathbf{\hat{P}}_{ari}$. As evidenced by Figure~\ref{fig-singular-ccr}, the cumulative contribution ratio of the first sixty (60/131) singular values ($ccr(60)$) exceeds $90\%$. This result indicates that the performance matrices exhibit a low-rank structure, which enables the use of low-rank matrix completion techniques~\cite{candes2012exact,fan2019factor, chi2018low} to reliably reconstruct the overall performance matrix from only a subset of observed performance entries.

To further verify the low-rank characteristics, we construct matrix completion tasks under the MCAR (missing completely at random) mechanism with missing rates $\in \{0.5, 0.6, 0.7, 0.8, 0.9\}$. We use matrix factorization and non-convex optimization techniques~\cite{chi2018low} to recover the missing entries of the performance matrices. Formally, we consider a rank-constrained least-squares problem:
\begin{equation}
    \underset{\bold{\Phi} \in \mathbb{R}^{m \times n}}{\min} ~\Vert \mathcal{P}_{\Omega} (\bold{\Phi} - \mathbf{\hat{P}}) \Vert_F^2, ~~~~\text{s.t.}~~\text{rank}(\bold{\Phi}) \leq r,
    \label{eq-1}
\end{equation}
where $\mathbf{\hat{P}} \in \mathbb{R}^{N \times H}$ denotes a specific performance matrix with $N$ rows (datasets) and $H$ columns (clustering algorithms with different HPCs), $\Omega$ consists of the locations of observed entries and the observation operator $\mathcal{P}_{\Omega}: \mathbb{R}^{N \times H} \rightarrow \mathbb{R}^{N \times H}$ as
\begin{equation}
    [\mathcal{P}_{\Omega}(\mathbf{\hat{P}})]_{ij} = 
        \begin{cases} 
            \mathbf{\hat{P}}_{ij}, & (i,j) \in \Omega \\
            0, & \text{otherwise}
        \end{cases}.
\end{equation}

Using the low-rank factorization $\bold{\Phi} = \mathbf{U}\mathbf{V}^T$, where $\mathbf{U} \in \mathbb{R}^{N \times r}$ and $\mathbf{V} \in \mathbb{R}^{H \times r}$, we can rewrite \eqref{eq-1} as an unconstrained optimization problem:
\begin{equation}
    \underset{\mathbf{U}, \mathbf{V}}{\min} ~\Vert \mathcal{P}_{\Omega} (\mathbf{U}\mathbf{V}^T - \mathbf{\hat{P}}) \Vert_F^2.
    \label{eq-2}
\end{equation}

In our experiments, we set $r=60$ and utilize the singular value decomposition technique to initialize the $\mathbf{U}=\mathbf{U}_0\bold{\Sigma}_0^{1/2}$ and $\mathbf{V}=\mathbf{V}_0\bold{\Sigma}_0^{1/2}$, where $\mathbf{U}_0\bold{\Sigma}_0\mathbf{V}_0^T$ is the best rank-$r$ approximation of $\mathcal{P}_{\Omega}(\mathbf{\hat{P}})$. The normalized Mean Absolute Percentage Error (MAPE) is employed to quantify the accuracy of the prediction.

\subsection{Algorithm Preference Analysis}
\label{appe-preference}

Beyond the overall performance analysis, we group the datasets from three perspectives to uncover more specific insights: (i) data modalities (image, text, tabular, bioinformatics). We treat bioinformatics datasets as a separate data category because they typically exhibit extremely high feature dimensionality, often with the number of features exceeding the number of samples. This characteristic distinguishes them substantially from other tabular datasets and leads to markedly different clustering behavior; (ii) feature dimensionality (low ($m\leq100$), middle ($100<m\leq500$), high ($m>500$)); and (iii) the degree of cluster imbalance, measured by IR (Section~\ref{appe-datasets}) (low ($\text{IR} < 0.1$), middle ($0.1 \leq \text{IR} \leq 0.3$), high ($\text{IR}>0.3$)).

\begin{figure}
    \centering
    \includegraphics[width=\linewidth]{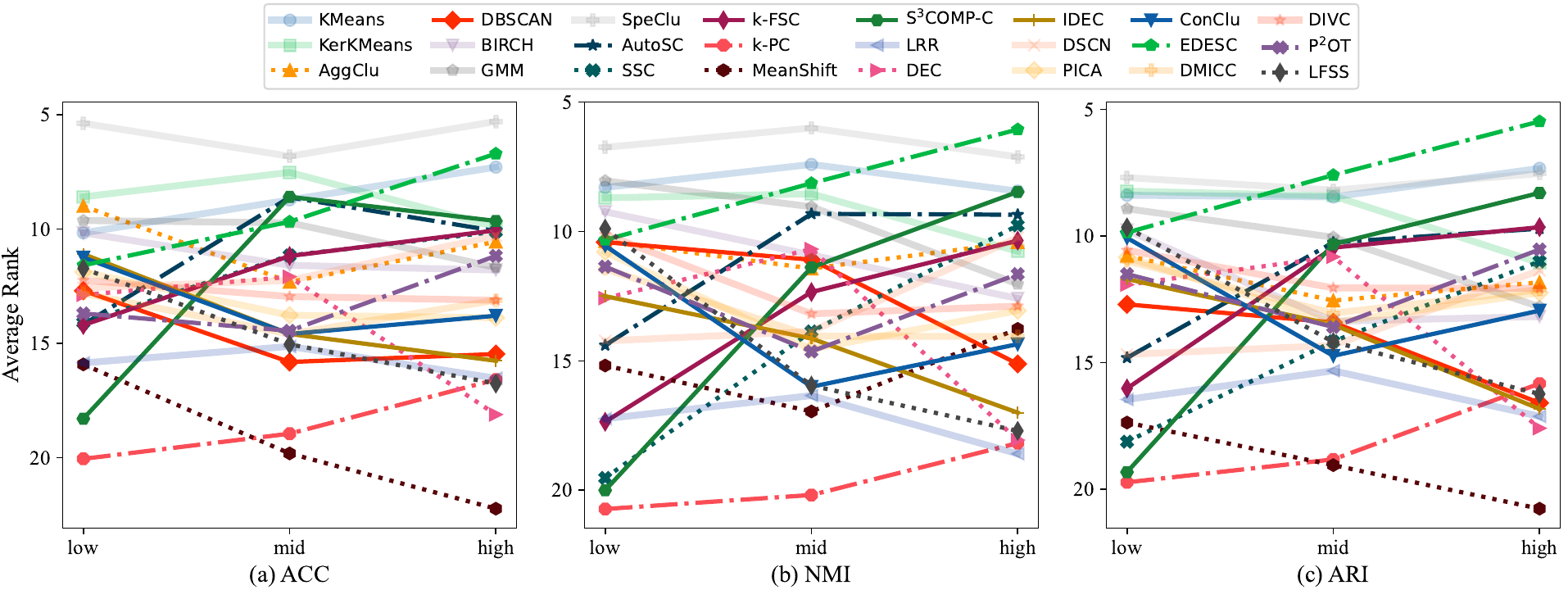}
    \caption{Algorithm preference analysis grouped by feature dimensionality under three evaluation metrics.}
    \label{fig-preference-dim}
\end{figure}

\begin{figure}
    \centering
    \includegraphics[width=\linewidth]{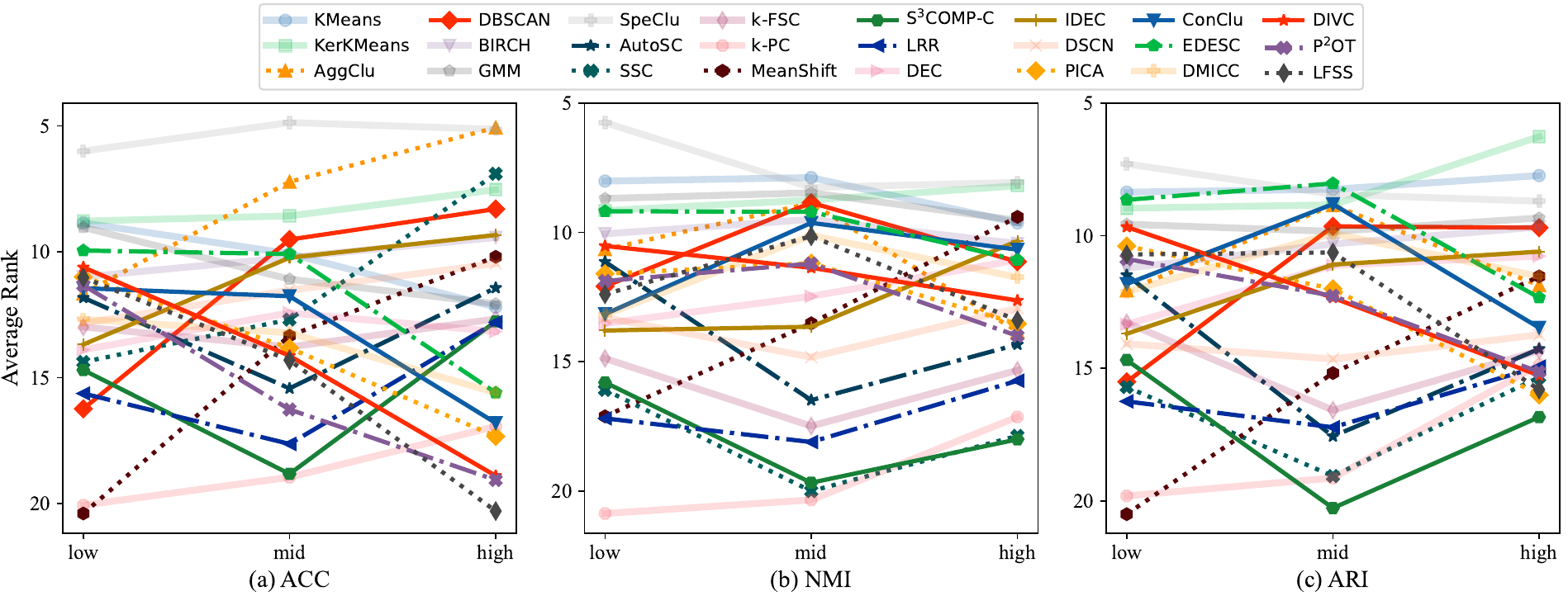}
    \caption{Algorithm preference analysis grouped by imbalance ratio under three evaluation metrics.}
    \label{fig-preference-IR}
\end{figure}

In Figure~\ref{fig-preference-type}, Figure~\ref{fig-preference-dim}, and Figure~\ref{fig-preference-IR}, we visualize the average ranks grouped by data type, feature dimensionality, and imbalance ratio, respectively. 
Although SpeClu consistently outperforms most methods in many scenarios, we also observe that certain methods are particularly effective in specific groups. For instance, AggClu demonstrates superior performance on highly imbalanced datasets, GMM is more effective on low-imbalance datasets, and EDESC outperforms other methods on high-dimensional datasets. On different data modalities, SpeClu shows consistent advantages compared with most baselines and AutoSC achieves superior performance on image data but inferior performance on other data types. Feature dimensionality, data modality, and cluster imbalance each induce distinct preference patterns, and algorithms with strong overall performance may still be suboptimal under specific dataset conditions (e.g., SpeClu). These findings underscore the importance of conditional algorithm selection rather than relying solely on global rankings.

\subsection{Model Selection Based on Performance Matrices}
\label{appe-ms}

\paragraph{Meta Features.}
First, various statistical characteristics of a dataset, such as the sample mean and higher-order moments, can be naturally regarded as meta-information. Following prior work~\cite{vanschoren2018meta, zhao2021automatic}, we extract 119 statistical variables as statistical meta-features. In addition to these statistical meta-features, we adopt KMeans as a landmark algorithm to compute KMeans-related meta-features. Specifically, we consider $K \in \{2, 4, 6, 8, 10, 12, 14, 16, 18, 20\}$, apply KMeans to each dataset for each value of $K$, and compute multiple internal evaluation metrics (IEM). These include four Silhouette-related statistics $SCs = \{\text{mean}, \text{std}, \text{min}, \text{max}\}$, the Calinski--Harabasz Index (CHI), the Davies--Bouldin Index (DBI), and seven sum-of-squared-error--related measures $SSEs = \{\text{total}, \text{mean}, \text{std}, \text{max}, \text{min}, \\ \text{explained\_ratio}, \text{unexplained\_ratio}\}$. In total, these yield 130 KMeans-based meta-features for each dataset. By concatenating the statistical and KMeans-based meta-features, we obtain a meta-feature vector $\mathbf{z}$ of dimensionality 249 for each dataset. Detailed information about these meta-features is provided in Table \ref{tab:meta-features}. 
\begin{table*}[t]
\centering
\caption{Meta-features for characterizing an arbitrary dataset.}
\label{tab:meta-features}
\resizebox{\textwidth}{!}{%
\begin{tabular}{llll}
\toprule
\textbf{Name} & \textbf{Formula} & \textbf{Rationale} & \textbf{Variants} \\
\midrule
Nr instances & $n$ & Speed, Scalability & $p/n$, $\log(n)$, $\log(n/p)$ \\
Nr features & $p$ & Curse of dimensionality & $\log(p)$, \% categorical \\
Sample mean & $\mu$ & Concentration & --- \\
Sample median & $\tilde{X}$ & Concentration & --- \\
Sample var & $\sigma^2$ & Dispersion & --- \\
Sample min & $\min(X)$ & Data range & --- \\
Sample max & $\max(X)$ & Data range & --- \\
Sample std & $\sigma$ & Dispersion & --- \\
Percentile & $P_i$ & Dispersion & $q1, q25, q75, q99$ \\
Interquartile Range (IQR) & $q_{75}-q_{25}$ & Dispersion & --- \\
Normalized mean & $\mu/\max(X)$ & Data range & --- \\
Normalized median & $\tilde{X}/\max(X)$ & Data range & --- \\
Sample range & $\max(X)-\min(X)$ & Data range & --- \\
Sample Gini & --- & Dispersion & --- \\
Median absolute deviation & $\mathrm{median}(|X-\tilde{X}|)$ & Variability and dispersion & --- \\
Average absolute deviation & $\mathrm{avg}(|X-\tilde{X}|)$ & Variability and dispersion & --- \\
Quantile Coefficient Dispersion & $\dfrac{q_{75}-q_{25}}{q_{75}+q_{25}}$ & Dispersion & --- \\
Coefficient of variance & --- & Dispersion & --- \\
Outlier outside 1\&99 & \% samples outside 1\% or 99\% & Basic outlying patterns & --- \\
Outlier 3STD & \% samples outside $3\sigma$ & Basic outlying patterns & --- \\
Normal test & If a sample differs from a normal dist. & Feature normality & --- \\
$k$th moments & --- & --- & 5th to 10th moments \\
Skewness & Feature skewness & Feature normality & min, max, $\mu$, $\sigma$, skewness, kurtosis \\
Kurtosis & $\mu_4/\sigma^4$ & Feature normality & min, max, $\mu$, $\sigma$, skewness, kurtosis \\
Correlation & $\rho$ & Feature interdependence & min, max, $\mu$, $\sigma$, skewness, kurtosis \\
Covariance & $\mathrm{Cov}$ & Feature interdependence & min, max, $\mu$, $\sigma$, skewness, kurtosis \\
Sparsity & $\#\text{Unique values}/n$ & Degree of discreteness & min, max, $\mu$, $\sigma$, skewness, kurtosis \\
ANOVA $p$-value & $p_{\text{ANOVA}}$ & Feature redundancy & min, max, $\mu$, $\sigma$, skewness, kurtosis \\
Coeff of variation & $\sigma_x/\mu_x$ & Dispersion & --- \\
Norm. entropy & $\dfrac{H(X)}{\log_2 n}$ & Feature informativeness & min, max, $\sigma$, $\mu$ \\
\midrule
Landmarker (KMeans) &--- & Cluster structure & 
SCs, CHI, DBI, SSEs \\

\bottomrule
\end{tabular}%
}
\end{table*}

\paragraph{Performance Matrices.}
Based on extensive experiments conducted on all benchmark datasets under all predefined hyperparameter configurations for each algorithm, we construct a performance vector $\hat{\mathbf{p}} \in \mathbb{R}^{H}$ for each dataset with respect to a given evaluation metric, where $H=273$ denotes the total number of clustering method hyperparameter combinations considered. By aggregating these performance vectors across all datasets, we obtain three performance matrices, namely $\hat{\mathbf{P}}_{\text{acc}}$, $\hat{\mathbf{P}}_{\text{nmi}}$, and $\hat{\mathbf{P}}_{\text{ari}} \in \mathbb{R}^{N \times H}$, corresponding to ACC, NMI, and ARI, respectively.

\paragraph{Model Selection.}
For each dataset $i$, we are given a meta-feature vector $\mathbf{z}_i$ and a corresponding performance vector $\mathbf{p}_i$ with respect to a specific evaluation metric. Our goal is to learn a mapping $f$ that effectively projects dataset meta-features into the performance space. This objective can be formulated as the following regression problem:
\begin{equation}
    \min_{f} \; \frac{1}{t} \sum_{i=1}^{t} \left\lVert f(\mathbf{z}_i) - \mathbf{p}_i \right\rVert^2 ,
\end{equation}
where $t$ denotes the number of training datasets.

To approximate the mapping $f$, we experiment with three regression models: XGBoost~\cite{chen2016xgboost}, LightGBM~\cite{ke2017lightgbm}, and Random Forest~\cite{breiman2001random}. All models are evaluated using 5-fold cross-validation for each evaluation metric. Given an unseen dataset with meta-feature vector $\mathbf{z}_{\text{new}}$, the trained regressor $f^*$ predicts the performance scores of all candidate models. The model corresponding to the highest predicted performance is then selected, namely,
\begin{equation}
    \arg\max_{j} \; f^*(\mathbf{z}_{\text{new}}).
\end{equation}

\section{Complete Best Performance (ACC, NMI, ARI) of 131 Benchmark Datasets}
\label{appe-complete-all-performance}
The detailed best performance of each clustering algorithm on 131 datasets is summarized in Table~\ref{per-beg} to Table~\ref{per-end}.

\begin{table*}[!ht]
\centering
\caption{The ACC (clustering accuracy) on datasets [1]-[44] (Part-1).}
\label{}
\resizebox{\textwidth}{!}{
\begin{tabular}{c|c|c|c|c|c|c|c|c|c|c|c|c}
\toprule
\textbf{Dataset Index} & KMeans & KerKMeans & AggClu & DBSCAN & BIRCH & GMM & SpeClu & AutoSC & SSC & k-FSC & k-PC & MeanShift \\
\midrule
1 & 0.885 & 0.885 & 0.902 & 0.885 & 0.918 & 0.803 & 0.918 & 0.803 & 0.738 & 0.587 & 0.659 & 0.770 \\
2 & 0.322 & 0.333 & 0.304 & 0.263 & 0.324 & 0.325 & 0.316 & 0.252 & 0.224 & 0.258 & 0.224 & 0.249 \\
3 & 0.972 & 0.963 & 0.972 & 0.956 & 0.968 & 0.947 & 0.971 & 0.635 & 0.659 & 0.704 & 0.655 & 0.908 \\
4 & 0.988 & 0.990 & 0.991 & 0.977 & 0.993 & 0.993 & 0.993 & 0.991 & 0.603 & 0.614 & 0.598 & 0.937 \\
5 & 0.547 & 0.547 & 0.430 & 0.519 & 0.542 & 0.446 & 0.462 & 0.449 & 0.355 & 0.450 & 0.358 & 0.472 \\
6 & 0.599 & 0.664 & 0.509 & 0.466 & 0.552 & 0.610 & 0.548 & 0.626 & 0.732 & 0.619 & 0.487 & 0.516 \\
7 & 0.555 & 0.588 & 0.720 & 0.703 & 0.577 & 0.709 & 0.725 & 0.692 & 0.720 & 0.696 & 0.715 & 0.714 \\
\bottomrule
\end{tabular}}
\label{per-beg}
\end{table*}

\begin{table*}[!ht]
\centering
\caption{The ACC (clustering accuracy) on datasets [1]-[44] (Part-2).}
\label{}
\resizebox{\textwidth}{!}{
\begin{tabular}{c|c|c|c|c|c|c|c|c|c|c|c|c}
\toprule
\textbf{Dataset Index} & S$^3$COMP-C & LRR & DEC & IDEC & DSCN & PICA & ConClu & EDESC & DMICC & DIVC & P$^2$OT & LFSS \\
\midrule
1 & 0.659 & 0.656 & 0.928 & 0.925 & 0.774 & 0.836 & 0.869 & 0.830 & 0.866 & 0.803 & 0.842 & 0.918 \\
2 & 0.246 & 0.257 & 0.372 & 0.326 & 0.257 & 0.300 & 0.354 & 0.316 & 0.320 & 0.296 & 0.299 & 0.283 \\
3 & 0.583 & 0.540 & 0.969 & 0.618 & 0.927 & 0.971 & 0.974 & 0.975 & 0.970 & 0.971 & 0.970 & 0.969 \\
4 & 0.569 & 0.603 & 0.991 & 0.990 & 0.904 & 0.993 & 0.992 & 0.956 & 0.974 & 0.993 & 0.922 & 0.975 \\
5 & 0.357 & 0.411 & 0.495 & 0.490 & 0.464 & 0.460 & 0.505 & 0.448 & 0.464 & 0.453 & 0.438 & 0.472 \\
6 & 0.573 & 0.660 & 0.590 & 0.618 & 0.420 & 0.658 & 0.526 & 0.544 & 0.609 & 0.646 & 0.631 & 0.616 \\
7 & 0.682 & 0.698 & 0.592 & 0.579 & 0.723 & 0.542 & 0.555 & 0.570 & 0.576 & 0.549 & 0.553 & 0.577 \\
\bottomrule
\end{tabular}}
\end{table*}

\begin{table*}[!ht]
\centering
\caption{The ACC (clustering accuracy) on datasets [45]-[88] (Part-1).}
\label{}
\resizebox{\textwidth}{!}{
\begin{tabular}{c|c|c|c|c|c|c|c|c|c|c|c|c}
\toprule
\textbf{Dataset Index} & KMeans & KerKMeans & AggClu & DBSCAN & BIRCH & GMM & SpeClu & AutoSC & SSC & k-FSC & k-PC & MeanShift \\
\midrule
45 & 0.324 & 0.333 & 0.386 & 0.303 & 0.343 & 0.310 & 0.382 & 0.244 & 0.316 & 0.332 & 0.238 & 0.330 \\
46 & 0.469 & 0.614 & 0.858 & 0.818 & 0.674 & 0.590 & 0.840 & 0.591 & 0.857 & 0.724 & 0.374 & 0.857 \\
47 & 0.556 & 0.565 & 0.701 & 0.763 & 0.706 & 0.629 & 0.649 & 0.529 & 0.657 & 0.589 & 0.621 & 0.617 \\
48 & 0.846 & 0.769 & 0.829 & 0.556 & 0.844 & 0.788 & 0.813 & 0.844 & 0.552 & 0.533 & 0.622 & 0.557 \\
49 & 0.609 & 0.612 & 0.584 & 0.571 & 0.664 & 0.630 & 0.598 & 0.521 & 0.596 & 0.596 & 0.529 & 0.593 \\
50 & 0.644 & 0.654 & 0.782 & 0.718 & 0.701 & 0.671 & 0.768 & 0.774 & 0.705 & 0.607 & 0.493 & 0.732 \\
51 & 0.964 & 0.846 & 0.849 & 0.707 & 0.749 & 0.908 & 0.964 & 0.955 & 0.927 & 0.915 & 0.587 & 0.721 \\
\bottomrule
\end{tabular}}
\end{table*}

\begin{table*}[!ht]
\centering
\caption{The ACC (clustering accuracy) on datasets [45]-[88] (Part-2).}
\label{}
\resizebox{\textwidth}{!}{
\begin{tabular}{c|c|c|c|c|c|c|c|c|c|c|c|c}
\toprule
\textbf{Dataset Index} & S$^3$COMP-C & LRR & DEC & IDEC & DSCN & PICA & ConClu & EDESC & DMICC & DIVC & P$^2$OT & LFSS \\
\midrule
45 & 0.263 & 0.285 & 0.303 & 0.308 & 0.327 & 0.247 & 0.318 & 0.330 & 0.257 & 0.254 & 0.249 & 0.274 \\
46 & 0.437 & 0.482 & 0.657 & 0.811 & 0.488 & 0.634 & 0.547 & 0.580 & 0.524 & 0.584 & 0.512 & 0.480 \\
47 & 0.530 & 0.587 & 0.676 & 0.723 & 0.557 & 0.577 & 0.576 & 0.609 & 0.643 & 0.639 & 0.587 & 0.620 \\
48 & 0.524 & 0.677 & 0.790 & 0.736 & 0.755 & 0.871 & 0.877 & 0.867 & 0.835 & 0.867 & 0.857 & 0.866 \\
49 & 0.528 & 0.529 & 0.630 & 0.640 & 0.612 & 0.600 & 0.603 & 0.595 & 0.614 & 0.609 & 0.607 & 0.677 \\
50 & 0.680 & 0.787 & 0.664 & 0.616 & 0.734 & 0.601 & 0.474 & 0.577 & 0.596 & 0.636 & 0.560 & 0.509 \\
51 & 0.879 & 0.863 & 0.856 & 0.803 & 0.820 & 0.798 & 0.704 & 0.830 & 0.836 & 0.690 & 0.894 & 0.849 \\
\bottomrule
\end{tabular}}
\end{table*}

\begin{table*}[!ht]
\centering
\caption{The ACC (clustering accuracy) on datasets [89]-[131] (Part-1).}
\label{}
\resizebox{\textwidth}{!}{
\begin{tabular}{c|c|c|c|c|c|c|c|c|c|c|c|c}
\toprule
\textbf{Dataset Index} & KMeans & KerKMeans & AggClu & DBSCAN & BIRCH & GMM & SpeClu & AutoSC & SSC & k-FSC & k-PC & MeanShift \\
\midrule
89 & 0.519 & 0.499 & 0.515 & 0.244 & 0.592 & 0.490 & 0.455 & 0.574 & 0.447 & 0.469 & 0.356 & 0.126 \\
90 & 0.251 & 0.261 & 0.230 & 0.160 & 0.227 & 0.243 & 0.241 & 0.301 & 0.217 & 0.205 & 0.212 & 0.220 \\
91 & 0.358 & 0.360 & 0.325 & 0.334 & 0.356 & 0.371 & 0.471 & 0.296 & 0.412 & 0.382 & 0.377 & 0.208 \\
92 & 0.665 & 0.832 & 0.575 & 0.667 & 0.575 & 0.727 & 0.817 & 0.575 & 0.383 & 0.482 & 0.303 & 0.402 \\
93 & 0.348 & 0.380 & 0.435 & 0.431 & 0.362 & 0.364 & 0.432 & 0.348 & 0.434 & 0.434 & 0.310 & 0.430 \\
94 & 0.855 & 0.912 & 0.924 & 0.915 & 0.837 & 0.852 & 0.814 & 0.531 & 0.837 & 0.812 & 0.832 & 0.892 \\
95 & 0.408 & 0.277 & 0.334 & 0.241 & 0.214 & 0.248 & 0.348 & 0.265 & 0.377 & 0.514 & 0.208 & 0.163 \\
\bottomrule
\end{tabular}}
\end{table*}

\begin{table*}[!ht]
\centering
\caption{The ACC (clustering accuracy) on datasets [89]-[131] (Part-2).}
\label{}
\resizebox{\textwidth}{!}{
\begin{tabular}{c|c|c|c|c|c|c|c|c|c|c|c|c}
\toprule
\textbf{Dataset Index} & S$^3$COMP-C & LRR & DEC & IDEC & DSCN & PICA & ConClu & EDESC & DMICC & DIVC & P$^2$OT & LFSS \\
\midrule
89 & 0.542 & 0.184 & 0.336 & 0.307 & 0.260 & 0.427 & 0.357 & 0.488 & 0.401 & 0.430 & 0.507 & 0.322 \\
90 & 0.252 & 0.222 & 0.200 & 0.217 & 0.237 & 0.199 & 0.213 & 0.210 & 0.211 & 0.214 & 0.195 & 0.187 \\
91 & 0.331 & 0.282 & 0.322 & 0.368 & 0.394 & 0.378 & 0.355 & 0.380 & 0.354 & 0.374 & 0.401 & 0.370 \\
92 & 0.443 & 0.528 & 0.615 & 0.638 & 0.626 & 0.932 & 0.732 & 0.601 & 0.809 & 0.938 & 0.698 & 0.887 \\
93 & 0.310 & 0.312 & 0.340 & 0.344 & 0.426 & 0.366 & 0.373 & 0.369 & 0.328 & 0.357 & 0.304 & 0.332 \\
94 & 0.738 & 0.607 & 0.776 & 0.856 & 0.846 & 0.752 & 0.778 & 0.734 & 0.796 & 0.696 & 0.642 & 0.711 \\
95 & 0.780 & 0.317 & 0.227 & 0.233 & 0.266 & 0.302 & 0.263 & 0.401 & 0.289 & 0.337 & 0.269 & 0.210 \\
\bottomrule
\end{tabular}}
\end{table*}

\begin{table*}[!ht]
\centering
\caption{The NMI on datasets [1]-[44] (Part-1).}
\label{}
\resizebox{\textwidth}{!}{
\begin{tabular}{c|c|c|c|c|c|c|c|c|c|c|c|c}
\toprule
\textbf{Dataset Index} & KMeans & KerKMeans & AggClu & DBSCAN & BIRCH & GMM & SpeClu & AutoSC & SSC & k-FSC & k-PC & MeanShift \\
\midrule
1 & 0.561 & 0.429 & 0.598 & 0.449 & 0.640 & 0.423 & 0.640 & 0.318 & 0.036 & 0.203 & 0.012 & 0.309 \\
2 & 0.143 & 0.152 & 0.134 & 0.118 & 0.128 & 0.131 & 0.128 & 0.073 & 0.014 & 0.059 & 0.036 & 0.075 \\
3 & 0.814 & 0.762 & 0.814 & 0.764 & 0.784 & 0.733 & 0.802 & 0.010 & 0.014 & 0.123 & 0.020 & 0.695 \\
4 & 0.907 & 0.918 & 0.925 & 0.882 & 0.939 & 0.939 & 0.939 & 0.924 & 0.001 & 0.038 & 0.018 & 0.763 \\
5 & 0.455 & 0.368 & 0.392 & 0.454 & 0.458 & 0.361 & 0.347 & 0.345 & 0.176 & 0.298 & 0.151 & 0.479 \\
6 & 0.633 & 0.630 & 0.459 & 0.558 & 0.623 & 0.612 & 0.566 & 0.622 & 0.642 & 0.524 & 0.398 & 0.526 \\
7 & 0.004 & 0.002 & 0.022 & 0.011 & 0.001 & 0.027 & 0.042 & 0.008 & 0.012 & 0.003 & 0.010 & 0.000 \\
\bottomrule
\end{tabular}}
\end{table*}

\begin{table*}[!ht]
\centering
\caption{The NMI on datasets [1]-[44] (Part-2).}
\label{}
\resizebox{\textwidth}{!}{
\begin{tabular}{c|c|c|c|c|c|c|c|c|c|c|c|c}
\toprule
\textbf{Dataset Index} & S$^3$COMP-C & LRR & DEC & IDEC & DSCN & PICA & ConClu & EDESC & DMICC & DIVC & P$^2$OT & LFSS \\
\midrule
1 & 0.013 & 0.036 & 0.645 & 0.639 & 0.193 & 0.442 & 0.414 & 0.403 & 0.415 & 0.312 & 0.456 & 0.640 \\
2 & 0.027 & 0.072 & 0.180 & 0.133 & 0.095 & 0.151 & 0.175 & 0.151 & 0.161 & 0.145 & 0.107 & 0.109 \\
3 & 0.025 & 0.024 & 0.789 & 0.617 & 0.653 & 0.803 & 0.834 & 0.832 & 0.804 & 0.803 & 0.813 & 0.815 \\
4 & 0.024 & 0.000 & 0.921 & 0.916 & 0.677 & 0.936 & 0.932 & 0.796 & 0.838 & 0.939 & 0.694 & 0.841 \\
5 & 0.141 & 0.270 & 0.335 & 0.338 & 0.340 & 0.346 & 0.417 & 0.326 & 0.347 & 0.382 & 0.355 & 0.396 \\
6 & 0.417 & 0.562 & 0.603 & 0.617 & 0.414 & 0.622 & 0.519 & 0.555 & 0.565 & 0.644 & 0.570 & 0.559 \\
7 & 0.005 & 0.000 & 0.001 & 0.002 & 0.030 & 0.006 & 0.002 & 0.003 & 0.004 & 0.007 & 0.001 & 0.013 \\
\bottomrule
\end{tabular}}
\end{table*}

\begin{table*}[!ht]
\centering
\caption{The NMI on datasets [45]-[88] (Part-1).}
\label{}
\resizebox{\textwidth}{!}{
\begin{tabular}{c|c|c|c|c|c|c|c|c|c|c|c|c}
\toprule
\textbf{Dataset Index} & KMeans & KerKMeans & AggClu & DBSCAN & BIRCH & GMM & SpeClu & AutoSC & SSC & k-FSC & k-PC & MeanShift \\
\midrule
45 & 0.097 & 0.110 & 0.087 & 0.095 & 0.088 & 0.097 & 0.076 & 0.070 & 0.022 & 0.009 & 0.050 & 0.056 \\
46 & 0.100 & 0.056 & 0.017 & 0.044 & 0.107 & 0.106 & 0.006 & 0.064 & 0.000 & 0.011 & 0.004 & 0.000 \\
47 & 0.003 & 0.070 & 0.120 & 0.185 & 0.131 & 0.027 & 0.001 & 0.001 & 0.029 & 0.010 & 0.016 & 0.101 \\
48 & 0.398 & 0.339 & 0.367 & 0.294 & 0.377 & 0.374 & 0.392 & 0.393 & 0.009 & 0.003 & 0.048 & 0.138 \\
49 & 0.031 & 0.030 & 0.021 & 0.059 & 0.060 & 0.038 & 0.003 & 0.002 & 0.001 & 0.000 & 0.002 & 0.002 \\
50 & 0.202 & 0.204 & 0.041 & 0.154 & 0.234 & 0.201 & 0.263 & 0.011 & 0.100 & 0.130 & 0.032 & 0.171 \\
51 & 0.939 & 0.906 & 0.905 & 0.791 & 0.869 & 0.903 & 0.927 & 0.910 & 0.831 & 0.817 & 0.518 & 0.812 \\
\bottomrule
\end{tabular}}
\end{table*}

\begin{table*}[!ht]
\centering
\caption{The NMI on datasets [45]-[88] (Part-2).}
\label{}
\resizebox{\textwidth}{!}{
\begin{tabular}{c|c|c|c|c|c|c|c|c|c|c|c|c}
\toprule
\textbf{Dataset Index} & S$^3$COMP-C & LRR & DEC & IDEC & DSCN & PICA & ConClu & EDESC & DMICC & DIVC & P$^2$OT & LFSS \\
\midrule
45 & 0.026 & 0.068 & 0.098 & 0.083 & 0.057 & 0.081 & 0.069 & 0.058 & 0.082 & 0.086 & 0.080 & 0.090 \\
46 & 0.001 & 0.046 & 0.149 & 0.136 & 0.088 & 0.060 & 0.118 & 0.054 & 0.086 & 0.068 & 0.084 & 0.081 \\
47 & 0.013 & 0.013 & 0.068 & 0.131 & 0.013 & 0.037 & 0.008 & 0.039 & 0.068 & 0.062 & 0.021 & 0.054 \\
48 & 0.003 & 0.094 & 0.335 & 0.256 & 0.295 & 0.451 & 0.463 & 0.439 & 0.399 & 0.438 & 0.415 & 0.436 \\
49 & 0.001 & 0.003 & 0.037 & 0.039 & 0.021 & 0.031 & 0.028 & 0.024 & 0.029 & 0.040 & 0.034 & 0.090 \\
50 & 0.114 & 0.099 & 0.191 & 0.167 & 0.069 & 0.099 & 0.135 & 0.143 & 0.166 & 0.066 & 0.133 & 0.161 \\
51 & 0.752 & 0.762 & 0.817 & 0.847 & 0.705 & 0.847 & 0.675 & 0.892 & 0.788 & 0.841 & 0.864 & 0.703 \\
\bottomrule
\end{tabular}}
\end{table*}

\begin{table*}[!ht]
\centering
\caption{The NMI on datasets [89]-[131] (Part-1).}
\label{}
\resizebox{\textwidth}{!}{
\begin{tabular}{c|c|c|c|c|c|c|c|c|c|c|c|c}
\toprule
\textbf{Dataset Index} & KMeans & KerKMeans & AggClu & DBSCAN & BIRCH & GMM & SpeClu & AutoSC & SSC & k-FSC & k-PC & MeanShift \\
\midrule
89 & 0.477 & 0.444 & 0.548 & 0.342 & 0.567 & 0.448 & 0.562 & 0.570 & 0.420 & 0.416 & 0.280 & 0.155 \\
90 & 0.074 & 0.080 & 0.015 & 0.177 & 0.015 & 0.018 & 0.026 & 0.074 & 0.026 & 0.040 & 0.036 & 0.104 \\
91 & 0.144 & 0.160 & 0.158 & 0.155 & 0.145 & 0.166 & 0.382 & 0.118 & 0.195 & 0.175 & 0.139 & 0.008 \\
92 & 0.772 & 0.761 & 0.801 & 0.852 & 0.801 & 0.788 & 0.797 & 0.728 & 0.266 & 0.581 & 0.346 & 0.678 \\
93 & 0.007 & 0.009 & 0.006 & 0.009 & 0.010 & 0.009 & 0.006 & 0.006 & 0.002 & 0.002 & 0.004 & 0.012 \\
94 & 0.187 & 0.435 & 0.543 & 0.483 & 0.000 & 0.158 & 0.373 & 0.000 & 0.000 & 0.112 & 0.000 & 0.292 \\
95 & 0.360 & 0.099 & 0.238 & 0.185 & 0.047 & 0.076 & 0.259 & 0.110 & 0.282 & 0.408 & 0.054 & 0.306 \\
\bottomrule
\end{tabular}}
\end{table*}

\begin{table*}[!ht]
\centering
\caption{The NMI on datasets [89]-[131] (Part-2).}
\label{}
\resizebox{\textwidth}{!}{
\begin{tabular}{c|c|c|c|c|c|c|c|c|c|c|c|c}
\toprule
\textbf{Dataset Index} & S$^3$COMP-C & LRR & DEC & IDEC & DSCN & PICA & ConClu & EDESC & DMICC & DIVC & P$^2$OT & LFSS \\
\midrule
89 & 0.546 & 0.070 & 0.219 & 0.206 & 0.267 & 0.356 & 0.271 & 0.446 & 0.327 & 0.391 & 0.430 & 0.175 \\
90 & 0.102 & 0.047 & 0.019 & 0.016 & 0.013 & 0.034 & 0.011 & 0.044 & 0.012 & 0.046 & 0.025 & 0.007 \\
91 & 0.188 & 0.033 & 0.069 & 0.149 & 0.164 & 0.152 & 0.179 & 0.166 & 0.102 & 0.188 & 0.178 & 0.120 \\
92 & 0.340 & 0.607 & 0.711 & 0.680 & 0.614 & 0.876 & 0.728 & 0.715 & 0.782 & 0.880 & 0.824 & 0.813 \\
93 & 0.001 & 0.004 & 0.006 & 0.003 & 0.007 & 0.004 & 0.012 & 0.008 & 0.006 & 0.003 & 0.009 & 0.009 \\
94 & 0.029 & 0.113 & 0.091 & 0.169 & 0.209 & 0.260 & 0.277 & 0.246 & 0.222 & 0.209 & 0.176 & 0.243 \\
95 & 0.645 & 0.219 & 0.061 & 0.064 & 0.117 & 0.197 & 0.182 & 0.289 & 0.158 & 0.243 & 0.136 & 0.064 \\
\bottomrule
\end{tabular}}
\end{table*}

\begin{table*}[!ht]
\centering
\caption{The ARI on datasets [1]-[44] (Part-1).}
\label{}
\resizebox{\textwidth}{!}{
\begin{tabular}{c|c|c|c|c|c|c|c|c|c|c|c|c}
\toprule
\textbf{Dataset Index} & KMeans & KerKMeans & AggClu & DBSCAN & BIRCH & GMM & SpeClu & AutoSC & SSC & k-FSC & k-PC & MeanShift \\
\midrule
1 & 0.585 & 0.575 & 0.636 & 0.589 & 0.690 & 0.398 & 0.690 & 0.358 & 0.051 & -0.009 & -0.001 & 0.330 \\
2 & 0.079 & 0.084 & 0.088 & 0.046 & 0.098 & 0.082 & 0.074 & 0.038 & 0.011 & 0.030 & 0.013 & 0.002 \\
3 & 0.891 & 0.858 & 0.891 & 0.831 & 0.874 & 0.799 & 0.885 & -0.013 & 0.015 & 0.165 & 0.043 & 0.815 \\
4 & 0.953 & 0.961 & 0.965 & 0.941 & 0.972 & 0.972 & 0.972 & 0.965 & -0.001 & 0.051 & 0.034 & 0.851 \\
5 & 0.284 & 0.241 & 0.199 & 0.302 & 0.282 & 0.190 & 0.220 & 0.225 & 0.107 & 0.205 & 0.098 & 0.288 \\
6 & 0.478 & 0.524 & 0.362 & 0.278 & 0.473 & 0.492 & 0.369 & 0.530 & 0.550 & 0.442 & 0.327 & 0.357 \\
7 & 0.005 & 0.010 & 0.016 & 0.045 & 0.002 & 0.084 & 0.033 & 0.045 & 0.016 & 0.012 & 0.005 & 0.000 \\
\bottomrule
\end{tabular}}
\end{table*}

\begin{table*}[!ht]
\centering
\caption{The ARI on datasets [1]-[44] (Part-2).}
\label{}
\resizebox{\textwidth}{!}{
\begin{tabular}{c|c|c|c|c|c|c|c|c|c|c|c|c}
\toprule
\textbf{Dataset Index} & S$^3$COMP-C & LRR & DEC & IDEC & DSCN & PICA & ConClu & EDESC & DMICC & DIVC & P$^2$OT & LFSS \\
\midrule
1 & -0.038 & 0.071 & 0.723 & 0.713 & 0.237 & 0.446 & 0.530 & 0.445 & 0.521 & 0.361 & 0.486 & 0.690 \\
2 & 0.019 & 0.036 & 0.108 & 0.087 & 0.041 & 0.085 & 0.123 & 0.093 & 0.104 & 0.081 & 0.069 & 0.079 \\
3 & 0.026 & -0.001 & 0.879 & 0.504 & 0.730 & 0.885 & 0.897 & 0.901 & 0.883 & 0.885 & 0.884 & 0.880 \\
4 & 0.019 & -0.000 & 0.962 & 0.958 & 0.727 & 0.971 & 0.969 & 0.842 & 0.900 & 0.972 & 0.727 & 0.903 \\
5 & 0.089 & 0.145 & 0.196 & 0.189 & 0.198 & 0.201 & 0.259 & 0.180 & 0.213 & 0.241 & 0.201 & 0.239 \\
6 & 0.349 & 0.492 & 0.480 & 0.504 & 0.215 & 0.519 & 0.385 & 0.422 & 0.459 & 0.521 & 0.466 & 0.468 \\
7 & 0.028 & 0.008 & 0.008 & 0.009 & 0.032 & 0.003 & 0.003 & 0.003 & 0.002 & 0.005 & -0.004 & 0.018 \\
\bottomrule
\end{tabular}}
\end{table*}

\begin{table*}[!ht]
\centering
\caption{The ARI on datasets [45]-[88] (Part-1).}
\label{}
\resizebox{\textwidth}{!}{
\begin{tabular}{c|c|c|c|c|c|c|c|c|c|c|c|c}
\toprule
\textbf{Dataset Index} & KMeans & KerKMeans & AggClu & DBSCAN & BIRCH & GMM & SpeClu & AutoSC & SSC & k-FSC & k-PC & MeanShift \\
\midrule
45 & 0.038 & 0.044 & 0.059 & 0.022 & 0.021 & 0.033 & 0.028 & 0.018 & 0.020 & -0.001 & 0.017 & 0.013 \\
46 & 0.048 & 0.058 & 0.041 & 0.067 & 0.108 & 0.050 & -0.024 & 0.088 & -0.002 & -0.046 & 0.002 & -0.001 \\
47 & 0.010 & 0.003 & 0.100 & 0.261 & 0.111 & 0.060 & 0.001 & 0.002 & 0.077 & 0.027 & 0.046 & 0.148 \\
48 & 0.477 & 0.289 & 0.432 & 0.278 & 0.474 & 0.331 & 0.392 & 0.474 & 0.010 & 0.003 & 0.059 & 0.055 \\
49 & 0.047 & 0.049 & 0.028 & 0.023 & 0.096 & 0.065 & 0.003 & 0.001 & 0.002 & 0.000 & 0.002 & 0.001 \\
50 & 0.188 & 0.200 & 0.041 & 0.188 & 0.266 & 0.231 & 0.392 & 0.006 & 0.226 & 0.152 & -0.039 & 0.181 \\
51 & 0.940 & 0.850 & 0.844 & 0.643 & 0.711 & 0.875 & 0.934 & 0.921 & 0.862 & 0.824 & 0.390 & 0.658 \\
\bottomrule
\end{tabular}}
\end{table*}

\begin{table*}[!ht]
\centering
\caption{The ARI on datasets [45]-[88] (Part-2).}
\label{}
\resizebox{\textwidth}{!}{
\begin{tabular}{c|c|c|c|c|c|c|c|c|c|c|c|c}
\toprule
\textbf{Dataset Index} & S$^3$COMP-C & LRR & DEC & IDEC & DSCN & PICA & ConClu & EDESC & DMICC & DIVC & P$^2$OT & LFSS \\
\midrule
45 & 0.003 & 0.026 & 0.031 & 0.022 & 0.003 & 0.024 & 0.036 & 0.028 & 0.025 & 0.022 & 0.020 & 0.028 \\
46 & -0.004 & 0.044 & 0.106 & 0.231 & 0.036 & 0.022 & 0.044 & -0.020 & 0.049 & 0.030 & 0.050 & 0.022 \\
47 & -0.000 & 0.028 & 0.097 & 0.153 & -0.012 & 0.041 & 0.020 & 0.056 & 0.094 & 0.088 & 0.033 & 0.058 \\
48 & 0.003 & 0.124 & 0.339 & 0.235 & 0.288 & 0.549 & 0.567 & 0.539 & 0.452 & 0.537 & 0.508 & 0.534 \\
49 & 0.003 & 0.003 & 0.065 & 0.072 & 0.040 & 0.041 & 0.042 & 0.037 & 0.054 & 0.053 & 0.045 & 0.124 \\
50 & 0.212 & 0.209 & 0.224 & 0.175 & 0.103 & 0.078 & 0.053 & 0.135 & 0.178 & 0.057 & 0.113 & 0.119 \\
51 & 0.785 & 0.701 & 0.782 & 0.747 & 0.677 & 0.753 & 0.670 & 0.830 & 0.768 & 0.687 & 0.861 & 0.701 \\
\bottomrule
\end{tabular}}
\end{table*}

\begin{table*}[!ht]
\centering
\caption{The ARI on datasets [89]-[131] (Part-1).}
\label{}
\resizebox{\textwidth}{!}{
\begin{tabular}{c|c|c|c|c|c|c|c|c|c|c|c|c}
\toprule
\textbf{Dataset Index} & KMeans & KerKMeans & AggClu & DBSCAN & BIRCH & GMM & SpeClu & AutoSC & SSC & k-FSC & k-PC & MeanShift \\
\midrule
89 & 0.345 & 0.321 & 0.400 & 0.057 & 0.429 & 0.322 & 0.314 & 0.443 & 0.279 & 0.299 & 0.175 & 0.003 \\
90 & 0.003 & 0.005 & 0.005 & 0.011 & -0.003 & 0.005 & 0.002 & 0.030 & -0.006 & 0.025 & 0.030 & 0.001 \\
91 & 0.109 & 0.102 & 0.083 & 0.100 & 0.090 & 0.097 & 0.278 & 0.055 & 0.174 & 0.134 & 0.128 & -0.000 \\
92 & 0.621 & 0.666 & 0.616 & 0.682 & 0.616 & 0.662 & 0.696 & 0.610 & 0.184 & 0.414 & 0.163 & 0.416 \\
93 & 0.006 & 0.011 & 0.002 & 0.008 & 0.023 & 0.020 & 0.000 & 0.007 & 0.001 & -0.000 & 0.004 & 0.005 \\
94 & 0.368 & 0.525 & 0.676 & 0.624 & 0.001 & 0.305 & 0.367 & 0.002 & -0.000 & 0.266 & -0.002 & 0.480 \\
95 & 0.223 & -0.010 & 0.057 & 0.012 & -0.009 & -0.001 & 0.131 & 0.007 & 0.173 & 0.317 & 0.017 & -0.002 \\
\bottomrule
\end{tabular}}
\end{table*}

\begin{table*}[!ht]
\centering
\caption{The ARI on datasets [89]-[131] (Part-2).}
\label{}
\resizebox{\textwidth}{!}{
\begin{tabular}{c|c|c|c|c|c|c|c|c|c|c|c|c}
\toprule
\textbf{Dataset Index} & S$^3$COMP-C & LRR & DEC & IDEC & DSCN & PICA & ConClu & EDESC & DMICC & DIVC & P$^2$OT & LFSS \\
\midrule
89 & 0.423 & 0.035 & 0.143 & 0.129 & 0.072 & 0.244 & 0.177 & 0.332 & 0.220 & 0.254 & 0.310 & 0.137 \\
90 & 0.055 & 0.011 & 0.009 & 0.013 & 0.002 & 0.022 & 0.009 & 0.030 & 0.009 & 0.029 & 0.015 & 0.005 \\
91 & 0.159 & 0.026 & 0.050 & 0.109 & 0.096 & 0.113 & 0.124 & 0.152 & 0.078 & 0.139 & 0.133 & 0.099 \\
92 & 0.281 & 0.503 & 0.542 & 0.565 & 0.474 & 0.851 & 0.588 & 0.561 & 0.694 & 0.860 & 0.671 & 0.770 \\
93 & 0.001 & 0.005 & 0.003 & 0.002 & 0.006 & 0.006 & 0.023 & 0.012 & 0.008 & 0.002 & 0.007 & 0.014 \\
94 & -0.083 & 0.043 & 0.210 & 0.269 & 0.268 & 0.244 & 0.287 & 0.235 & 0.293 & 0.150 & 0.085 & 0.173 \\
95 & 0.640 & 0.092 & -0.007 & 0.002 & 0.025 & 0.107 & 0.080 & 0.220 & 0.087 & 0.158 & 0.049 & 0.024 \\
\bottomrule
\end{tabular}}
\label{per-end}
\end{table*}

\end{document}